\documentclass[11pt]{article}

\usepackage[final]{acl}

\usepackage{times}
\usepackage{latexsym}
\usepackage[T1]{fontenc}
\usepackage[utf8]{inputenc}
\usepackage{microtype}
\usepackage{inconsolata}

\usepackage{graphicx}
\usepackage{booktabs}
\usepackage{amsmath, amsfonts, amssymb}
\usepackage{multirow}
\usepackage{multicol}
\usepackage{tablefootnote}
\usepackage{placeins}
\usepackage{afterpage}
\usepackage{enumitem}
\usepackage{url}
\usepackage{xurl} % allow URL line breaks at any character, fixes overfull hbox for long notion-style hash URLs
\usepackage{xcolor}
\usepackage{dsfont}
\usepackage{bbm}
\usepackage{tabularx}
\usepackage[table]{xcolor}
\usepackage{tikz}
\usetikzlibrary{arrows.meta, positioning, fit}
\usepackage{caption}
\usepackage{subcaption}

\definecolor{myDarkYellow}{rgb}{0.8, 0.6, 0.1}

\definecolor{myDarkGreen}{rgb}{0.1, 0.5, 0.2}

\definecolor{myDarkPurple}{rgb}{0.3, 0.1, 0.5}

\newcommand{\ieno}{\textit{i}.\textit{e}.}
\newcommand{\egno}{\textit{e}.\textit{g}.}

\definecolor{criticcolor}{RGB}{0,128,0}
\definecolor{criticcolor_0}{named}{red}
\newcommand{\scoreone}{\textcolor{green!50!black}{\texttt{<score>1</score>}}}
\newcommand{\scorezero}{\textcolor{red}{\texttt{<score>0</score>}}}

\newcommand{\authornote}[1]{%
  \begingroup
  \renewcommand{\thefootnote}{}%
  \footnotetext{#1}%
  \endgroup
}

\title{Stepwise Think-Critique: Interleaved Reasoning and Self-Critique \\ in a Single LLM}

\author{%
  Jiaqi Xu\textsuperscript{1}\textsuperscript{$\star$} \quad
  Cuiling Lan\textsuperscript{2} \quad
  Xuejin Chen\textsuperscript{1} \quad
  Yan Lu\textsuperscript{2} \\
  \normalfont\textsuperscript{1}\,University of Science and Technology of China \\
  \normalfont\textsuperscript{2}\,Microsoft Research Asia \\
  \normalfont\texttt{xujiaqi@mail.ustc.edu.cn}, \texttt{\{culan,yanlu\}@microsoft.com}, \texttt{xjchen99@ustc.edu.cn}
}

\begin{document}
\maketitle

\authornote{$\star$~This work was done when Jiaqi was an intern at MSRA. This paper is from an open collaborative research project.}

\begin{abstract}
Human beings solve complex problems through critical thinking, where reasoning and evaluation are intertwined to converge toward correct solutions.
However, most existing large language models (LLMs) treat the reasoning and verification as separate processes: they either generate reasoning without explicit self-checking or rely on external verifiers to detect errors post hoc. The former lacks immediate feedback, while the latter increases system complexity and hinders synchronized learning.
% ===== Backup of original abstract S3 (method paragraph) (replaced with single-LLM/inline framing per V4.0 reviewer pass) =====
% Motivated by human critical thinking, we propose \textbf{Stepwise Think-Critique (STC)}, a unified and end-to-end trainable framework that interleaves reasoning and self-critique at every intermediate step within a single model. STC is trained with a hybrid reinforcement learning objective that integrates reasoning rewards and critique-consistency rewards, thereby jointly optimizing solution correctness and reliability of self-evaluation.
% ===== End of backup =====
Motivated by human critical thinking, we propose \textbf{Stepwise Think-Critique (STC)}, an end-to-end trainable framework in which a single LLM emits a structured, step-level critique inline with each reasoning step. STC is trained with reinforcement learning that complements reasoning rewards with a critique-consistency reward derived from final-answer correctness, jointly optimizing reasoning correctness and critique reliability.
% ===== Backup of original abstract closing sentence (replaced with quantified version per V4.0 reviewer pass) =====
% Experiments on mathematical reasoning benchmarks show that STC demonstrates strong critical-thinking capabilities and produces more interpretable reasoning traces, representing a step toward \textbf{LLMs with built-in critical thinking}.
% ===== End of backup =====
On five mathematical reasoning benchmarks, STC improves Pass@1 by \textbf{7.2\%} over the 1.5B base model and
attains \textbf{67.4\%} step-level critique F1, surpassing seven external
process reward models evaluated at their per-dataset oracle
thresholds---a step toward
\textbf{LLMs with built-in critical thinking}.
\end{abstract}

% ===============================================================
\section{Introduction}

Large language models (LLMs) have demonstrated remarkable capabilities in solving complex reasoning tasks. Techniques such as Chain of Thought (CoT) prompting~\cite{wei2022chain} and reinforcement-based reasoning approaches (\egno, DeepSeek-R1)~\cite{guo2025deepseek} enable models to decompose problems into multi-step reasoning trajectories, leading to improved performance in domains such as mathematical problem-solving, program synthesis, and complex question-answering \cite{guo2025deepseek,yu2025dapo,chen2025towards}. 

Despite these advances, a fundamental limitation persists: current approaches lack an intrinsic, on-the-fly mechanism to assess the correctness of intermediate reasoning steps. Such a mechanism is highly desirable because i) in domains such as mathematics, science, and finance, users require transparent judgments on whether each reasoning step is correct; ii) integrated self-critique allows models to adapt their reasoning trajectories dynamically, enabling self-assessment during inference; and iii) stepwise evaluation could provide dense feedback, facilitating more effective policy optimization.

\begin{figure*}[t]
    \centering
    %\vspace{-3mm}
    \includegraphics[width=0.82\linewidth]%
    {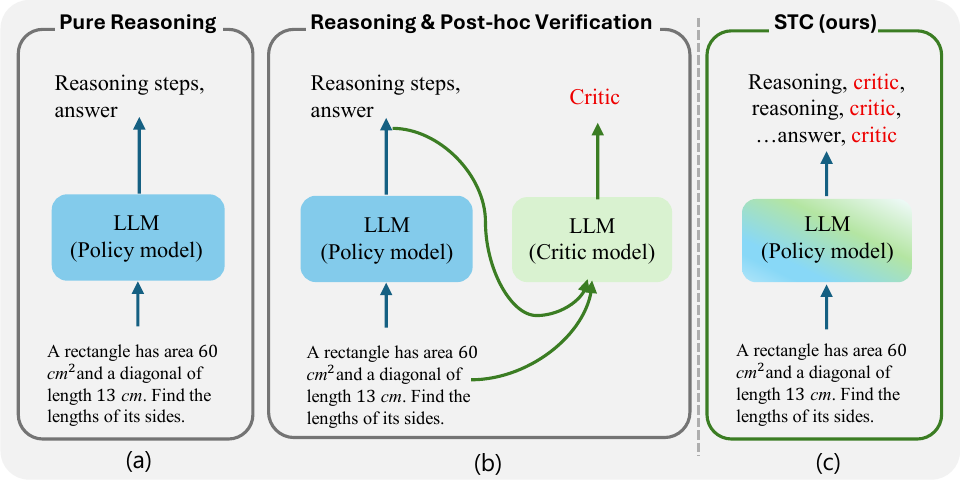}
    %{figures/illustration.pdf}
    %\vspace{-5mm}
    \caption{
        \textbf{Paradigms of reasoning and verification in LLMs.} 
        (a) \emph{Pure reasoning}: a model generates multi-step reasoning and answers without assessment. 
        (b) \emph{Post-hoc verification}: reasoning is produced first and then critiqued by an external verifier (\egno, process reward model), which provides delayed feedback and increases system complexity. 
        (c) \emph{Stepwise Think-Critique (\textbf{ours})}: a single model outputs interleaved reasoning and critique at each step, enabling timely self-assessment and improved interpretability.
        }
    \label{fig:illustration}
    \vspace{-2mm}
\end{figure*}

Most existing reasoning models \cite{guo2025deepseek,openai_o3_2025} focus solely on generating reasoning and final answers, as illustrated in Figure~\ref{fig:illustration}~(a). Some studies have explored post-hoc verification, where an external verifier evaluates the correctness of the reasoning process (see Figure~\ref{fig:illustration}~(b)). For example, process reward models (PRMs) have been developed to evaluate intermediate reasoning steps produced by a policy model~\cite{zhang2025lessons,zou2025reasonflux,khalifa2025process,xiong2025stepwiser}. Although such methods can detect errors, the post-hoc nature (\ieno, the separation of reasoning and verification) prevents timely feedback, and maintaining separate reasoning policy and verifier models increases system complexity, hindering seamless deployment in real-world pipelines.\footnote{In deployment, passing trajectories between models requires additional scheduling, memory management, and fault tolerance burdens, which increases system fragility.}

%Human beings solve complex problems through critical thinking, continuously interleaving reasoning and evaluation (judgement).
The ability to interleave reasoning and evaluation is a hallmark of human problem solving. Humans naturally engage in critical thinking, continuously reflecting on intermediate conclusions, assessing their validity, and adjusting subsequent steps accordingly. This process allows humans not only to reach correct solutions more efficiently but also to produce reasoning traces that are interpretable and trustworthy.
Inspired by this, we propose a novel framework called Stepwise Think-Critique (STC). As illustrated in Figure~\ref{fig:illustration}~(c), unlike existing paradigms that separate reasoning and verification, our method unifies them. At each step of problem solving, the model not only generates the next reasoning step but also the justification for its correctness. 
We instantiate our framework within an LLM that interleaves reasoning and self-critique steps. The model is trained with RL using a hybrid objective that combines reasoning rewards (based on the final outcomes), critique-consistency rewards (encouraging the model’s self-assessment over the final answer to align with the ground-truth correctness), and format rewards for structured outputs. STC endows the model with reasoning and critiquing capabilities.

\setlist[itemize]{leftmargin=0.6em,topsep=0.0em,itemsep=0.0em}
Our work makes the following contributions:
\begin{itemize}
    % ===== Backup of original Bullet 1 (replaced with single-LLM/inline framing per V4.0 reviewer pass) =====
    % \item We propose \textbf{Stepwise Think-Critique}, a unified framework that imitates human critical thinking by interleaving reasoning and self-critique at each step of problem solving.
    % ===== End of backup =====
    \item We propose \textbf{Stepwise Think-Critique (STC)}, a framework
   in which a single LLM emits a structured, step-level critique \emph{inline with} each
   reasoning step, eliminating the post-hoc reasoner--verifier split.
   % ===== Backup of original Bullet 2 (replaced with end-to-end RL/masking framing per V4.0 reviewer pass) =====
   % \item We design a hybrid reinforcement learning objective that combines critique-consistency rewards to guide the critic’s reasoning quality, and reasoning rewards to optimize the policy’s reasoning performance.
   % ===== End of backup =====
   \item We design an \textbf{end-to-end RL training scheme} that
   jointly optimizes reasoning and critique within a single
   model: a critique-consistency reward derived
   from final-answer correctness is propagated only to critique
   tokens via masking, for accurate credit assignment.
   %Critic’s step-wise feedbacks are used for dense reward optimization.
   % ===== Backup of original Bullet 3 (replaced with quantified empirical claims per V4.0 reviewer pass) =====
   % \item Experiments show that \ourss demonstrates strong critical-thinking capabilities and provides more interpretable traces.
   % ===== End of backup =====
   \item On five math benchmarks, STC improves Pass@1 by
   \textbf{7.2\%} and attains \textbf{67.4\%} step-level critique F1 on average,
   outperforming seven external process reward models evaluated at
   their per-dataset oracle thresholds.
   % \item Experiments demonstrate that our framework significantly improves performance on mathematical reasoning tasks, while also providing  interpretability.
\end{itemize}
Our unified framework eliminates the requirement of an external verifier and enables on-the-fly self-critique.
This represents a crucial step towards \textbf{building LLMs with built-in critical thinking}, bridging the gap between human-like problem solving and reliable machine reasoning. 
At inference time, we can optionally enable the critiquing for enhanced interpretability—providing judgments on where and why mistakes occur—or disable it to reduce token usage. 

%During inference, the critique can be optionally enabled to support critical reasoning or disabled to reduce token usage.

\begin{figure*}[!t]
    \centering
    %\resizebox{0.95\linewidth}{!}{
    \includegraphics[width=0.82\linewidth]{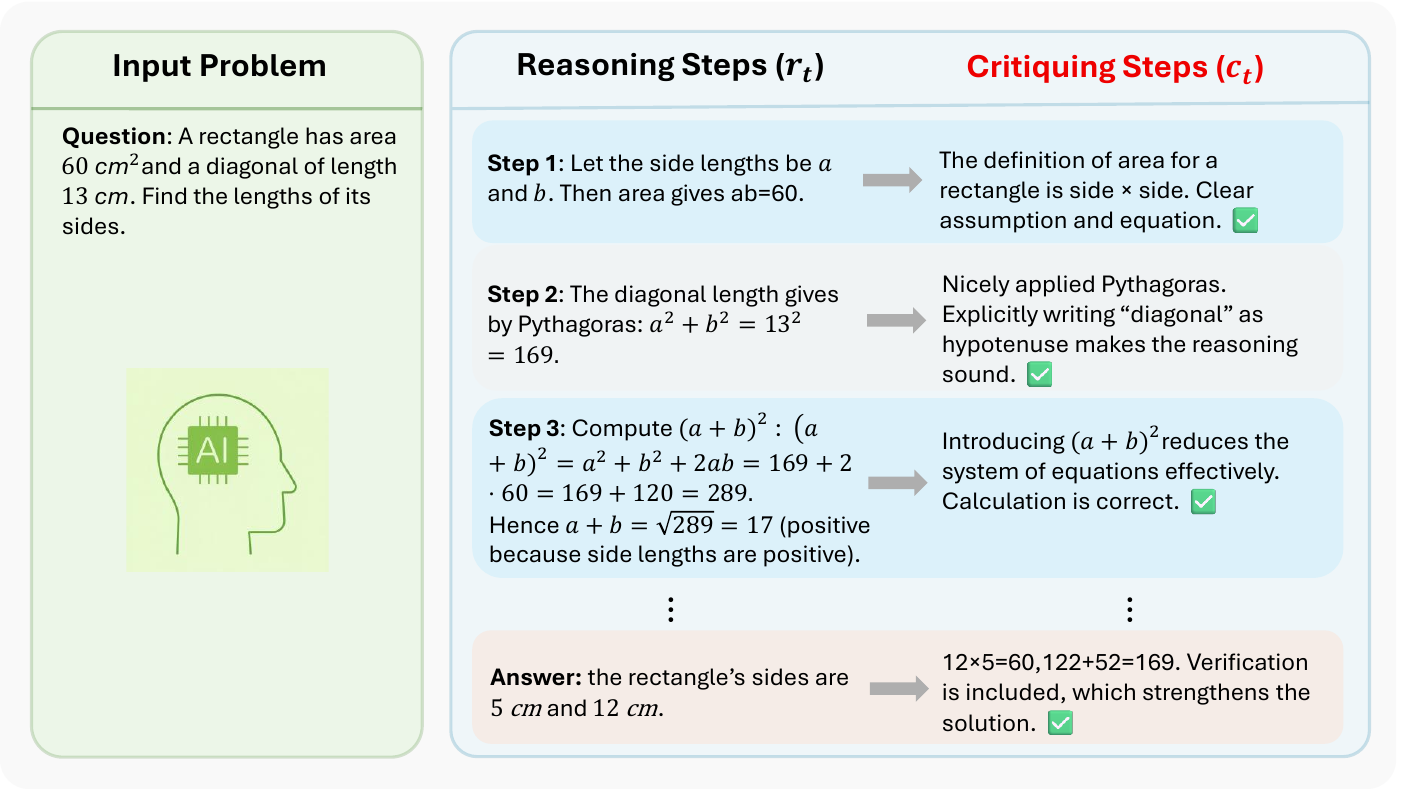}
    %}
    \caption{
        Illustration of \textbf{Stepwise Think-Critique (STC).}
        Given an input problem, the model alternates between generating a reasoning step ($r_t$) and a critiquing step ($c_t$).
        Critiques provide immediate correctness assessment, which supports on-the-fly error localization.
        At inference time, users may choose \emph{compact mode} (reasoning only) or \emph{full mode} (reasoning+critique) depending on efficiency vs.\ interpretability requirements.
    }
    \label{fig:approach}
    \vspace{-2mm}
\end{figure*}

\section{Related Work}

\subsection{RL for LLM Reasoning}

Reinforcement learning (RL) has become a central paradigm for eliciting multi-step reasoning---``thinking''---in LLMs, with works such as DeepSeek-R1~\cite{guo2025deepseek}, DAPO~\cite{yu2025dapo}, and REINFORCE++~\cite{hu2025reinforce++} reinforcing sequences of reasoning actions that lead to correct solutions.

% ===== Backup of §2.1 Para 2 (compressed below; intro/abstract already cover the human-critical-thinking motivation) =====
% However, human problem-solving relies not only on stepwise reasoning but also on critical evaluation: self-reflection, error detection, and iterative correction. Some RL approaches, such as actor–critic methods like VAPO~\cite{yue2025vapo}, employ a value model (critic) to estimate the expected return of actions or tokens. While effective for reward optimization, these critics do not explicitly assess the correctness of reasoning steps or provide interpretable feedback. Existing RL frameworks lack intrinsic, fine-grained self-assessment capabilities, limiting their potential for trustworthy and transparent problem-solving.
% ===== End of backup =====
Actor--critic methods like VAPO~\cite{yue2025vapo} use a value model to estimate expected return but do not explicitly assess the correctness of reasoning steps or provide interpretable feedback. Existing RL frameworks thus lack intrinsic, fine-grained self-assessment capabilities.

A separate line of work studies LLM \emph{self-correction}, where an explicit refinement pass consumes a response-level critique signal---Self-Refine~\cite{madaan2023selfrefine}, Self-Verification~\cite{weng2023selfverify}, CRITIC~\cite{gou2024critic} (tool-augmented), and SCoRe~\cite{kumar2024score}. DeepSeek-R1~\cite{guo2025deepseek} shows reflection-style ``wait'' tokens can emerge mid-trajectory from outcome-only RL, while such reflection is unstructured and unsupervised.

\emph{STC pursues a different objective}: equipping a single model with an intrinsic, consistency-supervised, step-level critique capability, directly enabling error localization and interpretability without an external verifier.

\subsection{Process Reward Modelling}

Process reward models (PRMs) provide another perspective on stepwise evaluation, focusing on the intermediate reasoning steps rather than solely on the final answer~\cite{zhong2025comprehensive}. By offering finer-grained feedback, PRMs could improve credit assignment, help identify local errors.

Designs of PRMs have evolved along multiple axes. Early approaches adopt a discriminative format, where a classification head predicts step-level correctness scores~\cite{lightman2023let,zhang2025lessons}. More recent works embrace generative approaches, producing natural language judgments for each reasoning step~\cite{zhang2024generative,zheng2025processbench,khalifa2025process,zhao2025genprm}. ThinkPRM~\cite{khalifa2025process}, for instance, generates a verification chain-of-thought (CoT) that critiques and justifies each step post hoc, improving interpretability and error localization.

Beyond scoring individual steps, PRMs are also applied during policy optimization, where they provide step-level reward signals that guide model updates. Despite their utility, most PRMs are trained offline on curated datasets and remain frozen, which can make them brittle under distribution shifts. PRIME~\cite{cui2025process} mitigates this limitation by deriving implicit process rewards from outcome labels through SFT, while TANGO~\cite{zha2025rl} jointly trains a generator and PRM via RL. Nevertheless, these methods rely on separate models, which adds complexity, communication overhead, and memory usage. \emph{In contrast, STC integrates the critic into a single model, avoiding extra communication and deployment overhead.}

% ===== Backup of §2.2 Para 4 (deleted; STC positioning already covered in §2.1's v6 paragraph) =====
% In contrast, our work equips a \emph{single} model with both reasoning and intrinsic stepwise self-assessment. This integrated approach facilitates co-evolution of the reasoner and critic during RL, allows on-the-fly self-critique of each reasoning step without requiring additional models, simplifying deployment, improving efficiency and transparency.
% ===== End of backup =====
%In contrast, our work equips a \emph{single} model with both reasoning and intrinsic stepwise self-assessment, facilitating on-the-fly assessment without increasing system communication complexity.

\section{Stepwise Think-Critique (STC)}

\subsection{Overview of STC}

We introduce \textbf{Stepwise Think-Critique (STC)}, a unified framework that interleaves reasoning and critique within an output sequence. Unlike prior paradigms that either perform pure reasoning or rely on external post-hoc verification, STC equips a single model with the ability to generate reasoning steps and simultaneously assess their correctness.  

Formally, given an input problem $x$, the model generates a sequence that alternates between reasoning step $\{r_t\}_{t=1}^T$ and critique step $\{c_t\}_{t=1}^T$, where $T$ denotes the number of steps that varies for different sequences. As illustrated in Figure~\ref{fig:approach}, each reasoning step $r_t$ advances the solution, while the corresponding critique $c_t$ provides a natural-language judgment and score for assessing the correctness of $r_t$. 
The final solution is produced in the last reasoning step, followed by its associated critique. 
%The critiques serve as intermediate rewards during training and as interpretability signals in inference. 

We first train the model with supervised fine-tuning (SFT) on our synthesized reasoning–critique data, enabling it to learn the interleaved reasoning-critique patterns and to acquire basic critique ability (for cold start). We then further train the model with RL using reasoning reward,  critique-consistency reward, and format reward, to boost critical thinking capability.

\subsection{Interleaved Reasoning-Critiquing}
\label{subsec:interleave}

For an LLM, to enable interleaved generation, we design instruction formats that explicitly encourage the model to alternate between reasoning and critique. Specifically, the prompt specifies the required structure:  
\begin{equation}
   r_1 \rightarrow c_1 \rightarrow r_2 \rightarrow c_2 \rightarrow \cdots \rightarrow r_T \rightarrow c_T.
\end{equation}
Reasoning steps are partitioned according to the generated $\texttt{\textbackslash n\textbackslash n}$. Each critique step is expressed in natural language and consists of two parts: 
(i) a concise justification, and 
(ii) a binary correctness label, where $1$ denotes ``Correct'' and $0$ denotes ``Incorrect.'' 
The model outputs critiques using the following tagged format:
\texttt{<critic>}\{\emph{justification}\}\texttt{</critic>} and a binary judgment in the format \texttt{<score>}\{\emph{1 or 0}\}\texttt{</score>}.
%\(\langle \texttt{score}\rangle\) \{\emph{1 or 0}\} \(\langle/\texttt{score}\rangle\). 
This structured format allows the model to explain errors in context while keeping the output compact and informative.  

During inference, through prompting alone, we support two modes:  
\begin{itemize}
    \item \textbf{Compact mode}: reasoning only, omitting critiques to save tokens.
    \item \textbf{Full mode}: reasoning and critique are interleaved, providing detailed step-level judgment, enabling error localization and enhancing interpretability.  
\end{itemize}  

This dual-mode design offers a flexible trade-off between efficiency and interpretability, allowing users to select the mode that best suits their needs.

%\subsection{Training}

\subsection{Supervised Fine-Tuning (SFT)}

We begin with supervised fine-tuning (SFT) to equip the model with interleaved reasoning-critique generation. We synthesize SFT data by sampling reasoning trajectories $(r_1,\ldots,r_T)$ from the base model on training problems, then prompting a strong LLM (\egno, GPT-5) to produce stepwise critiques $c_n = (j_n, s_n)$ following the format in \S\ref{subsec:interleave}, where $j_n$ is a natural-language justification and $s_n \in \{0,1\}$ a binary correctness label. To remove unreliable supervision, a rule-based verifier $V$ checks the final answer $r_T$, and we discard trajectories where the strong LLM's score on $r_T$ disagrees with $V$. We then fine-tune the model on the resulting $\mathcal{D} = \{(x, \tau)\}$ with $\tau = (r_1, c_1, \ldots, r_T, c_T)$ by minimizing $\mathcal{L}_{\text{SFT}}(\theta) = -\mathbb{E}_{(x,\tau)\sim \mathcal{D}} \log \pi_\theta(\tau \mid x)$.

\subsection{RL with Joint Reasoning and Critiquing}

While SFT provides initial alignment, it is limited by data coverage and generalization.  
We therefore further post-train the model using Grouped Reinforcement Policy Optimization (GRPO) with designed rewards.  

\subsubsection{Reward Design}
\label{subsub:reward_design}

%\noindent\textbf{Rewarding Design.} 

We define rewards at three levels:  

\noindent\emph{\textbf{Reasoning reward}:} We assign reward based on whether the answer is correct or not: 
\begin{equation}
R_{\text{reason}}(\tau) = \mathbbm{1}[r_T = y],
\label{eq:reasoning-reward}
\end{equation}
where $y$ is the ground-truth answer. $\mathbbm{1}[\cdot]$ denotes indicator function. 

\noindent\emph{\textbf{Critique-consistency reward (Critique reward)}:}  
Since no step-level ground-truth critiques are available during training, we derive supervision from the correctness of the final answer.  
Let $z \in \{0,1\}$ be the true correctness indicator of the final answer $r_T$, \ieno,
\begin{equation}
    z = \mathbbm{1}[r_T = y].
\end{equation}
Recall that $s_T$ denotes the binary label predicted by the critique for the final answer (\ieno, reasoning step $r_T$). We then define the critique reward w.r.t. the final answer as
\begin{equation}
    R_{\text{crit}}(c_T) = \mathbbm{1}[s_T = z],
    \label{eq:critique-reward}
\end{equation}
which \emph{encourages the model to produce critiques consistent with the actual correctness of the final answer}. 
\emph{Similar to the RL mechanism that the final reasoning reward can drive the learning of intermediate reasoning steps~\cite{guo2025deepseek}, our critique reward drives the learning of intermediate critic steps, which naturally penalizes false judgments that do not lead to correct final critique.}

\noindent\emph{\textbf{Format reward}:}  
To ensure structural validity, we require that each reasoning step is followed by a well-formed critique (\ieno, both justification and binary label).  
Let $v_n \in \{0,1\}$ be an indicator of whether the $n$-th reasoning--critique pair is correctly formatted.  
The format reward is defined as the fraction of valid pairs among all $T$ reasoning steps:
\begin{equation}
    R_{\text{format}}(\tau) = \frac{1}{T} \sum_{n=1}^{T} v_n,
\end{equation}
where $R_{\text{format}}(\tau) \in [0,1]$ encourages the model to consistently produce structurally correct interleaved outputs.
%encourages the model to maintain consistently valid interleaved outputs.

\subsubsection{Optimization}
%\paragraph{Optimization.}

We leverage GRPO to update model $\pi_\theta$ to maximize expected advantages. For an input problem $x$, the policy model samples a group of $G$ individual trajectories $\{\tau_k\}_{k=1}^G$. 
\emph{To ensure correct credit assignment, we perform separate optimization for the critique parts.}

Specifically, the critique advantage for the $k$-th trajectory is computed by normalizing the group-level critique-consistency rewards:  
\begin{equation}
    A_{\text{crit}}^{(k)} =
    \frac{
        R_{\text{crit}}^{(k)} -
        \mathrm{mean}\!\left(\{R_{\text{crit}}^{(k)}\}_{k=1}^G\right)
    }{
        \mathrm{std}\!\left(\{R_{\text{crit}}^{(k)}\}_{k=1}^G\right)
    }.
\label{eq:A_critic}
\end{equation}

The gradients from critique-consistency rewards are only back-propagated to \textbf{critique tokens} within $\{c_n^{(k)}|n = 1, 2, \cdots, T\}$ for accurate credit assignment. Therefore, the critique-consistency advantage for the $t$-th token in $\tau_k$ is
\begin{equation}
    \mathcal{A}_{\text{crit}}^{k,t} = A_{\text{crit}}^{(k)} \cdot \mathbbm{1}\!\left[t \in \text{Critique}(\tau_k)\right],
\end{equation}
where $t \in \text{Critique}(\tau_k)$ denotes that the $t$-th token belongs to critique tokens.

Considering the critiques would influence the next step reasoning, the reasoning reward and format reward are used to optimize all the tokens. Similar to (\ref{eq:A_critic}), we obtain reasoning advantage $A_{\text{reason}}^{(k)}$ and format advantage $A_{\text{format}}^{(k)}$, respectively.

The combined advantage is then:
\begin{equation}
    \mathcal{A}_{k,t} = \lambda_\text{crit} \alpha \mathcal{A}_{\text{crit}}^{k,t} + \lambda_\text{reason} A_{\text{reason}}^{(k)} + \lambda_\text{format} A_{\text{format}}^{(k)},
\label{eq:joint-adv}
\end{equation}
where $\lambda_\text{crit}$, $\lambda_\text{reason}$, $\lambda_\text{format}$ control contributions; $\alpha = \sum_k|\tau_k| / \sum_k|\tau_k^{\text{crit}}|$ undoes the dilution from GRPO's token-level averaging (Eq.~\ref{eq:grpo-objective} in Appendix~A), since $\mathcal{A}_{\text{crit}}^{k,t}$ is supported on only $|\tau_k^{\text{crit}}|$ tokens.

We optimize $\mathcal{A}_{k,t}$ via DAPO-style token-level GRPO~\cite{yu2025dapo} with a KL constraint to a reference policy (Appendix~\ref{sec:grpo_full}, Eq.~\ref{eq:grpo-objective}).

This joint objective encourages accurate reasoning, faithful critiques, and well-structured outputs, enabling the model not only to solve problems but also to verify and explain its reasoning process.

\section{Experiments}

\begin{table*}[!t]
\centering
\scriptsize
\setlength{\tabcolsep}{4pt}
\renewcommand{\arraystretch}{1.05}
\begin{tabular*}{\linewidth}{l@{\extracolsep{\fill}}*{6}{cc}}
\toprule
\multirow{2}{*}{\textbf{Model}}
& \multicolumn{2}{c}{\textbf{AIME24}} & \multicolumn{2}{c}{\textbf{AMC23}}
& \multicolumn{2}{c}{\textbf{MATH-500}} & \multicolumn{2}{c}{\textbf{Minerva}}
& \multicolumn{2}{c}{\textbf{OlympiadBench}} & \multicolumn{2}{c}{\textbf{Average}} \\
\cmidrule(lr){2-3}\cmidrule(lr){4-5}\cmidrule(lr){6-7}\cmidrule(lr){8-9}\cmidrule(lr){10-11}\cmidrule(lr){12-13}
& P@1 & P@8 & P@1 & P@8 & P@1 & P@8 & P@1 & P@8 & P@1 & P@8 & P@1 & P@8 \\
\midrule
DS-Qwen-1.5B & 17.5 & 43.3 & 51.1 & 72.3 & 76.2 & 89.6 & 25.1 & 46.3 & 36.0 & 52.2 & 41.2 & 60.7 \\
\midrule
STC-SFT & 18.3 & 46.7 & 49.1 & 71.1 & 73.3 & 88.6 & 20.3 & 39.7 & 34.2 & 49.3 & 39.1 & 59.1 \\
\addlinespace[2pt]
STC (compact) & \textbf{27.9} & \textbf{63.3} & \textbf{62.7} & \textbf{81.9} & 82.3 & \textbf{93.4} & 27.8 & 48.5 & 41.8 & 57.2 & \textbf{48.5} & \textbf{68.9} \\
\addlinespace[2pt]
STC (full) & \textbf{27.9} & 56.7 & 60.5 & \textbf{81.9} & \textbf{82.9} & 92.6 & \textbf{27.9} & \textbf{49.3} & \textbf{42.7} & \textbf{57.3} & 48.4 & 67.6 \\
%STC-GRPO w/o Dense reward  & 27.9  & 56.7  & 60.5  & 81.9  & 82.9  & 92.6  & 27.9  & 49.3  & 42.7  & 57.3  & 48.4  & 67.6 \\
%STC-GRPO (compact) & 25.8  & 56.7  & 63.3  & 80.7  & 82.6  & 93.8  & 27.7  & 49.3  & 42.2  & 56.3  & 48.3  & 67.4 \\
%STC-GRPO (full) & 30.4 & 53.3 & 64.2 & 83.1 & 82.6 & 92.0 & 27.3 & 48.2 & 43.6 & 59.0 & 49.6 & 67.1
\bottomrule
\end{tabular*}
\caption{Reasoning performance (\%) of models on math reasoning benchmarks, evaluated with Pass@1 (P@1) and Pass@8 (P@8). \textbf{Bold} marks the best per column.}
\label{tab:main-acc}
%\vspace{-6mm}
\end{table*}

% ===== Table 2 source moved here (right after Table 1) to encourage p.6 placement =====
\begin{table*}[!t]%[htbp]
\centering
\scriptsize
\setlength{\tabcolsep}{3pt}
\renewcommand{\arraystretch}{1.0}
\begin{tabular*}{\linewidth}{l@{\extracolsep{\fill}}ccc ccc ccc ccc ccc ccc}
\toprule
\multirow{2}{*}{Model}
& \multicolumn{3}{c}{\textbf{AIME24}}
& \multicolumn{3}{c}{\textbf{AMC23}}
& \multicolumn{3}{c}{\textbf{MATH-500}}
& \multicolumn{3}{c}{\textbf{Minerva}}
& \multicolumn{3}{c}{\textbf{OlympiadBench}}
& \multicolumn{3}{c}{\textbf{Average}} \\
\cmidrule(lr){2-4}
\cmidrule(lr){5-7}
\cmidrule(lr){8-10}
\cmidrule(lr){11-13}
\cmidrule(lr){14-16}
\cmidrule(lr){17-19}
& Corr. & Err. & \textbf{F1} &
Corr. & Err. & \textbf{F1} &
Corr. & Err. & \textbf{F1} &
Corr. & Err. & \textbf{F1} &
Corr. & Err. & \textbf{F1} &
Corr. & Err. & \textbf{F1} \\
\midrule

STC-SFT
& 100 & 10.5 & 19.0
& 98.5 & 22.6 & 36.8
& 99.2 & 7.4 & 13.8
& 96.6 & 8.7 & 16.0
& 98.5 & 7.8 & 14.4
& 98.8 & 11.4 & 20.0 \\
\addlinespace[2pt]

STC w/o Critic
& 97.1 & 25.7 & 40.7
& 95.2 & 26.7 & 41.7
& 97.4 & 16.6 & 28.4
& 94.6 & 12.7 & 22.4
& 94.2 & 24.2 & 38.5
& 95.7 & 21.2 & 34.3 \\
\addlinespace[2pt]

STC
& 82.1 & 60.3 & \textbf{69.6}
& 90.3 & 55.1 & \textbf{68.4}
& 96.2 & 32.6 & \textbf{48.7}
& 90.1 & 21.9 & \textbf{35.2}
& 90.5 & 43.1 & \textbf{58.4}
& 89.8 & 42.6 & \textbf{56.1} \\

\bottomrule
\end{tabular*}
\captionof{table}{
\textbf{Answer-level critic} performance (\%) evaluated against \textbf{verifiable ground truth}.
Corr. denotes correct-answer accuracy ($\uparrow$).
Err. denotes error-answer accuracy ($\uparrow$).
F1 ($\uparrow$), as the overall evaluation, is the harmonic mean of Corr. and Err.
}
\label{tab:ans_critic_results_wo_VR}
%\vspace{-8mm}
\end{table*}

\subsection{Experimental Setup}

\subsubsection{SFT Data Construction}
% ===== Backup of original §4.1.1 (uncompressed, 2 bullets with detailed stats) =====
% We construct the SFT dataset by sampling from the training pool and generating stepwise critiques with GPT-5. We then apply filtering to ensure paired structure and answer-level consistency.
% \begin{itemize}
%     \item \textbf{Scale \& Filtering:} We randomly select 10k samples and generate one reasoning rollout per sample using the base model. We then filter out samples with reasoning token length larger than 8,192, and use GPT-5 to generate critiques for each reasoning step. Next, we apply a three-stage filter to ensure: (i) structural integrity (reasoning and critique blocks are strictly paired); (ii) answer--critique consistency with the ground-truth critic. This yields \textbf{5,056} high-quality samples (about 51\% retention). For each retained sample, we interleave reasoning and critique steps to construct a unified trace.
%     \item \textbf{Diversity \& Depth:} The dataset spans a wide range of complexities with token lengths up to 15,130. About 75\% of samples fall between 2,048 and 8,192 tokens, 24\% exceed 8,192 tokens, and 10\% exceed 10,240 tokens, providing sufficient depth for complex multi-step reasoning.
%     %\item \textbf{Noise Reduction:} This rigorous filtering (especially the outcome-consistency check) ensures critiques are not merely stylistic but anchored to verifiable final answers.
% \end{itemize}
% ===== End of backup =====

We construct the SFT dataset by sampling 10k base-model rollouts on the training pool, generating stepwise critiques with GPT-5, and filtering for structural integrity and answer-critique consistency, yielding \textbf{5{,}056} interleaved samples ($\sim$51\% retention). Length statistics and the full filtering pipeline are detailed in Appendix~\ref{sec:sft_data_details}.

\subsubsection{Implementation Details}
% ===== Backup of original §4.1.2 (full hyperparameter listing moved to Appendix~\ref{sec:impl_setup}) =====
% We use DeepSeek-R1-Distill-Qwen-1.5B (abbreviated as DS-Qwen-1.5B)~\cite{guo2025deepseek} as our 1.5B-parameter base model. We take
% DeepScaleR-Preview-Dataset \cite{deepscaler2025} as our training dataset. SFT is performed with a batch size of 64 for 550 steps. For RL with GRPO, we set the group size to $G=8$, the batch size to 32, and $\lambda_\text{format}=0.05$. The system prompt is shown in Table~\ref{tab:qualitative} (in orange). We train RL for about 1200 steps. The maximum response lengths are set to 8,192 tokens for reasoning and 4,096 tokens for critique. More details are provided in Appendix.
% ===== End of backup =====
We use DS-Qwen-1.5B~\cite{guo2025deepseek} as the 1.5B-parameter base model and train on DeepScaleR-Preview-Dataset~\cite{deepscaler2025}. Hyperparameters, max response lengths, and the system prompt are in Appendix~\ref{sec:impl_setup}.

\subsubsection{Evaluation Protocol} 
We evaluate both the reasoning and critique capability of models. Evaluations are performed with a maximum response length of 8,192 tokens for reasoning and 4,096 tokens for critique by default. 

For reasoning quality, we report \textbf{Pass@1} (P@1 for short) and \textbf{Pass@8}.

For critique quality, we evaluate both the \textbf{answer level} and the \textbf{process level}. At the answer level, correctness is directly \textbf{verifiable} from the final solution (see Eq.~\ref{eq:critique-reward}).
At the process level, because step-level human labels are unavailable, we use a strong LLM judge (GPT-5-mini) to produce pseudo labels for stepwise critiques.

We report three metrics: \textbf{Correct Accuracy (Corr.)}, \textbf{Error Accuracy (Err.)}, and their harmonic mean \textbf{F1}; higher is better for all ($\uparrow$). 
Corr. measures how well the critic accepts truly correct items, Err. measures how well it rejects truly incorrect items.
Following \cite{zheng2025processbench}, we \textbf{use F1 to assess the overall critic capability}. This metric penalizes degenerate solutions that achieve high correct accuracy but fail to detect errors, or vice versa.
Detailed metric definitions (including step-level and answer-level) are provided in Appendix~\ref{sec:eval_metrics}.

\subsubsection{Evaluation Datasets} We evaluate our framework on multiple mathematical reasoning benchmarks: \textbf{AIME24}, \textbf{MATH-500}, \textbf{AMC23}, \textbf{Minerva}, and \textbf{OlympiadBench}. These datasets vary in problem complexity and problem styles, providing a comprehensive testbed.

\subsection{Results}

We evaluate STC on three axes: intrinsic reasoning quality (\S\ref{subsub:reasoning}), critique quality (\S\ref{subsub:critique}), and critique quality relative to seven external PRMs (\S\ref{subsub:prm_compare}). For the first two axes we compare three model variants: \emph{DS-Qwen-1.5B} (base model), \emph{STC-SFT} (after supervised fine-tuning), and \emph{STC} (RL-trained). \S\ref{subsub:qual} shows qualitative trajectories.

\subsubsection{Reasoning Performance}
\label{subsub:reasoning}

Table~\ref{tab:main-acc} reports reasoning accuracy across all benchmarks.
\textbf{i)}~SFT reduces average Pass@1 accuracy by 2.1\% relative to the base model. We attribute this drop to enabling stepwise critique, which shifts the model’s output distribution and temporarily degrades reasoning performance.
\textbf{ii)}~Subsequent RL training recovers and surpasses this gap: \emph{STC (full)} outperforms \emph{STC-SFT} by \textbf{9.3\%} at Pass@1 and \textbf{8.5\%} at Pass@8 on average, and also exceeds the base DS-Qwen-1.5B by \textbf{7.2\%} Pass@1, showing that the inline-critic objective is compatible with strong reasoning.

\paragraph{Full mode vs.\ compact mode.}
STC can operate in two inference modes: \emph{compact mode}, where critique generation is disabled and the model outputs only reasoning, and \emph{full mode}, where the model performs both reasoning and stepwise critique. The two modes share the same model weights and differ \emph{only} in the system prompt (highlighted in \textcolor{orange}{orange} in Tables~\ref{tab:qualitative} and~\ref{tab:example-compact}). As shown in Table~\ref{tab:main-acc}, both modes outperform STC-SFT significantly and achieve comparable reasoning accuracy: since the reasoning capability is acquired during training, toggling critique generation at inference time does not affect it. Users can flexibly choose the mode based on whether intermediate critiques are needed, without sacrificing reasoning performance.

% ===== Table 3 source moved here (between §4.2.1 and §4.2.2) to target p.7 placement =====
\begin{table*}[!t]%[htbp]
\centering
\scriptsize
\setlength{\tabcolsep}{3pt}
\renewcommand{\arraystretch}{1.0}
\begin{tabular*}{\linewidth}{l@{\extracolsep{\fill}}ccc ccc ccc ccc ccc ccc}
\toprule
\multirow{2}{*}{Model}
& \multicolumn{3}{c}{\textbf{AIME24}}
& \multicolumn{3}{c}{\textbf{AMC23}}
& \multicolumn{3}{c}{\textbf{MATH-500}}
& \multicolumn{3}{c}{\textbf{Minerva}}
& \multicolumn{3}{c}{\textbf{OlympiadBench}}
& \multicolumn{3}{c}{\textbf{Average}} \\
\cmidrule(lr){2-4}
\cmidrule(lr){5-7}
\cmidrule(lr){8-10}
\cmidrule(lr){11-13}
\cmidrule(lr){14-16}
\cmidrule(lr){17-19}
& Corr. & Err. & \textbf{F1} &
Corr. & Err. & \textbf{F1} &
Corr. & Err. & \textbf{F1} &
Corr. & Err. & \textbf{F1} &
Corr. & Err. & \textbf{F1} &
Corr. & Err. & \textbf{F1} \\
\midrule

STC-SFT
& 93.6 & 35.4 & 51.4
& 94.5 & 31.3 & 47.1
& 97.3 & 24.2 & 38.8
& 95.4 & 19.3 & 32.2
& 90.6 & 45.0 & 60.1
& 94.3 & 31.1 & 45.9 \\
\addlinespace[2pt]

STC w/o Critic
& 94.0 & 35.2 & 51.2
& 94.8 & 30.0 & 45.6
& 96.1 & 25.0 & 39.7
& 94.6 & 20.4 & 33.5
& 89.9 & 47.3 & 61.9
& 93.9 & 31.6 & 46.4 \\
\addlinespace[2pt]

STC
& 84.4 & 61.3 & \textbf{71.0}
& 84.0 & 59.5 & \textbf{69.7}
& 89.7 & 53.3 & \textbf{66.9}
& 89.6 & 37.3 & \textbf{52.7}
& 77.9 & 75.3 & \textbf{76.6}
& 85.1 & 57.3 & \textbf{67.4} \\

\bottomrule
\end{tabular*}
\captionof{table}{
\textbf{Process-level critic} performance (\%) evaluated by GPT-5-mini.
To save cost, we tested on the first 30 prompts $\times$ 8 rollouts $=$ 240 trajectories/samples for each dataset.
}
\label{tab:process_critic_results}
\end{table*}

\subsubsection{Critique Performance}
\label{subsub:critique}

We evaluate critique quality using Corr.\ (accuracy on truly correct instances), Err.\ (accuracy on truly incorrect instances), and their harmonic mean F1 at both answer-level (Table~\ref{tab:ans_critic_results_wo_VR}) and step-level (Table~\ref{tab:process_critic_results}). We compare our final model \emph{STC} with two variants: \emph{STC-SFT} (SFT only), and \emph{STC w/o Critic} (RL without critique reward).

\paragraph{Answer-level critique (Table~\ref{tab:ans_critic_results_wo_VR}).}
\textbf{i)} \emph{STC-SFT} exhibits a strong positive bias: it achieves 98.8\% Corr.\ but only 11.4\% Err., indicating poor error detection.  Table~\ref{tab:answer_level_confusion} in Appendix~\ref{sec:more_stats} shows that only ~4\% of answers are predicted as incorrect, while 27\% are actually incorrect from the valid samples.
\textbf{ii)} Adding RL with only answer and format rewards (\ieno, \emph{STC w/o Critic}) modestly increases Err.\ to 21.2\%, but overall F1 remains low.
\textbf{iii)} Incorporating the critique-consistency reward (\ieno, \emph{STC}) substantially improves error detection, \textbf{raising Err. from 21.2\% to 42.6\%}. Although Corr. drops about 6\%, \textbf{the overall F1 improves markedly from 34.3\% to 56.1\%}, indicating a much better balance between recognizing correct and incorrect answers.
Table~\ref{tab:answer_level_confusion} in Appendix~\ref{sec:more_stats} shows that the predicted negative ratio increases from 4\% (\emph{STC-SFT}) to 21\% (\emph{STC w/ Critic}), approaching the ground-truth ratio of 29\%.

% ===== Tables (Step-level Critic) ----------------

\paragraph{Step-level critique (Table~\ref{tab:process_critic_results}).}
A similar trend is observed for the step-level critic.
\textbf{i)} \emph{STC-SFT} shows a strong positive bias (94.3\% Corr., 31.1\% Err.). Table~\ref{tab:process_level_confusion} in Appendix~\ref{sec:more_stats} shows that only 11\% of steps are predicted as erroneous, despite a ground-truth error rate of about 20\%.
\emph{STC w/o Critic} behaves similarly.
\textbf{ii)} With the critique reward, \emph{STC} \textbf{substantially improves error detection, raising Err. from 31.6\% to 57.3\%} and \textbf{F1 from 46.4\% to 67.4\%}, revealing that our hybrid rewards effectively encourage balanced and reliable step-level critique.
Table~\ref{tab:process_level_confusion} in Appendix~\ref{sec:more_stats} shows that the predicted error-step ratio increases from 11\% to 23\%, approaching the ground-truth ratio of 19\%.

\paragraph{Reliability of the GPT-5-mini judge.}
Since human-annotated step-level labels are unavailable, we employ GPT-5-mini as an automatic judge for process-level evaluation, which may introduce noise.
To assess its reliability, we compare GPT-5-mini judgments with verifiable answer-level ground truth on a subset, obtaining an agreement rate of about \textbf{90\%}. This suggests that GPT-5-mini provides reasonably reliable pseudo ground truth for stepwise critique evaluation.

To further validate the reliableness of GPT-5-mini step labels, we use a stronger (more expensive) LLM, Claude Opus 4.7 (with maximum reasoning budget and no truncation), as a reference and re-evaluate the AIME24 labeled corpus. Treating Claude as the gold standard, GPT-5-mini achieves a balanced \textbf{F1 of 87.3\%} (Corr.\ 88.3\%, Err.\ 86.4\%, raw agreement 88.1\%). To balance performance and cost, we use GPT-5-mini as the judge. More analysis can be found in Appendix~\ref{sec:claude_reliability}.

\paragraph{Stepwise critique as a dense reward.}
We also explored stepwise critiques as a dense RL reward; the formulation we tried did not yield robust gains (see Appendix~\ref{sec:dense_reward_discussion}).

% ===== Table 4 source moved here (after Figure 3) so Fig 3 appears above Tab 4 visually =====
\begin{table*}[!t]
\centering
\scriptsize
\setlength{\tabcolsep}{3pt}
\renewcommand{\arraystretch}{1.0}
\begin{tabular*}{\linewidth}{l@{\extracolsep{\fill}}cccccc}
\toprule
\textbf{Method} & \textbf{AIME24} & \textbf{AMC23} & \textbf{MATH-500} & \textbf{Minerva} & \textbf{OlympiadBench} & \textbf{Average} \\
\midrule
Math-Shepherd-7B~\cite{wang2024math}       & 66.3 & 64.8 & 70.8 & 61.8 & 61.4 & 65.0 \\
RLHFlow-PRM-8B~\cite{dong2024rlhf}         & 58.6 & 58.4 & 66.5 & 58.7 & 58.9 & 60.2 \\
Skywork-PRM-1.5B~\cite{skywork-o1-prm}     & 51.0 & 53.4 & 68.3 & 53.7 & 49.5 & 55.2 \\
Skywork-PRM-7B~\cite{skywork-o1-prm}       & 47.7 & 52.8 & 68.1 & 54.3 & 46.8 & 53.9 \\
Qwen-Math-PRM-7B~\cite{zhang2025lessons}        & 62.4 & 66.0 & \textbf{74.7} & \textbf{64.5} & 61.9 & 65.9 \\
PathFinder-PRM-7B~\cite{pala2025pathfinder}& 61.5 & 60.9 & 69.1 & 57.3 & 56.7 & 61.1 \\
Granite-PRM-8B~\cite{ibm2026granite}       & 59.7 & 59.9 & 66.6 & 59.9 & 56.9 & 60.6 \\
\midrule
\textbf{STC} (1.5B, calib-free) & \textbf{71.0} & \textbf{69.7} & 66.9 & 52.7 & \textbf{76.6} & \textbf{67.4} \\
\bottomrule
\end{tabular*}
\caption{Step-level critique F1 (\%) comparison of STC and external PRMs. Each external PRM cell reports F1 at its per-dataset \textbf{oracle} threshold; STC is calibration-free with a single binary prediction per step.}
\label{tab:prm_compare}
\end{table*}

\subsubsection{Comparison with External Process Reward Models}
\label{subsub:prm_compare}

We compare STC against seven PRMs: Math-Shepherd-Mistral-7B~\cite{wang2024math}, RLHFlow-PRM-Deepseek-8B~\cite{dong2024rlhf}, Skywork-o1-PRM-1.5B/7B~\cite{skywork-o1-prm}, Qwen2.5-Math-PRM-7B~\cite{zhang2025lessons}, PathFinder-PRM-7B~\cite{pala2025pathfinder}, and Granite-3.3-8B Math PRM v2~\cite{ibm2026granite}.

All systems are evaluated on the same $1{,}200$ STC reasoning trajectories (by dropping critics) (30 prompts $\times$ 8 rollouts $\times$ 5 datasets) against the same GPT-5-mini pseudo-GT step labels. Each PRM emits a continuous score per step; we sweep $\sim$200 binarization thresholds in $[0.001, 0.999]$ and report the per-dataset \textbf{oracle} threshold (F1-maximizing -- an upper bound requiring labeled tuning per dataset). STC is calibration-free, emitting a single binary score per step.

% Figure relocated to Appendix to save body-paper space (fig:prm_corr_err_curves_step, fig:prm_corr_err_curves_answer).

Table~\ref{tab:prm_compare} shows the step-level performance comparison. STC reaches \textbf{67.4\%} average step-F1, ahead of the strongest PRM at its oracle threshold (Qwen-Math-PRM-7B, $65.9\%$), and is best on 3 of 5 benchmarks (AIME24, AMC23, OlympiadBench) -- despite being a 1.5B model with no per-dataset threshold tuning. Figure~\ref{fig:prm_corr_err_curves_step} in Appendix~\ref{sec:prm_answer_level} shows the full Corr.\ vs.\ Err.\ curves, where STC achieves the best performance on average. Beyond accuracy, STC is also more efficient: it saves $\sim$28\%/$\sim$75--77\% FLOPs and $\sim$49\%/$\sim$82--84\% peak GPU memory versus the 1.5B/7B--8B PRM pipelines (Appendix~\ref{sec:prm_answer_level}, Table~\ref{tab:inference_cost}).

\paragraph{Note on ProcessBench compatibility.}
We attempted ProcessBench~\cite{zheng2025processbench} via post-hoc verification prompts, but the 1.5B base model is too small to reliably emit ProcessBench's required error-index format; ProcessBench compatibility likely requires scaling to a larger backbone, which we leave for future work.

\subsubsection{Qualitative Analysis}
\label{subsub:qual}

\textbf{Full mode:} Some typical full-mode trajectories from STC are shown in Tables~\ref{tab:qualitative} to~\ref{tab:example_olympiad_primes} in Appendix~\ref{sec:more_qualitative}. Critiques are informative yet concise: they provide explicit error causes for incorrect steps and brief confirmations for correct steps. Table~\ref{tab:qualitative} shows STC catching a wrong intermediate guess ($1/\binom{6}{2}$). Table~\ref{tab:example_math_army} shows STC identifying the cause of an error (a misread problem constraint).

\noindent\textbf{Compact mode:} The same problems in compact mode (reasoning only) are shown in Tables~\ref{tab:example-compact} to~\ref{tab:example-compact-olympiad} in Appendix~\ref{sec:compact_mode}.

\noindent\textbf{Failures:} STC's critic is not always reliable. Table~\ref{tab:wrong_example_minerva} in Appendix~\ref{sec:failure_cases} shows a trajectory where STC missed a wrong unit conversion ($3\,\mathrm{Mpc}\to3000\,\mathrm{pc}$) at step 3.

\section{Conclusion}

We presented Stepwise Think-Critique (STC), a unified framework that interleaves reasoning and self-critique within a single model.
Across multiple mathematical reasoning benchmarks, STC demonstrates strong reasoning and critiquing capabilities and produces more interpretable process traces. These results support STC as a practical step toward LLMs with built-in critical thinking, bridging the gap between human-like problem solving and reliable machine reasoning.

We hope this work motivates further research on critical-thinking mechanisms that jointly improve transparency, reliability, and reasoning quality.

% ===============================================================
\section{Limitations}

While STC demonstrates promising results on mathematical
reasoning, several limitations remain.

\paragraph{Scale and single-seed evaluation.}
We validate the framework on a single 1.5B-parameter base model
(DS-Qwen-1.5B) under a fixed random seed; multi-seed experiments
were not run due to compute cost (one full SFT+RL run requires
${\sim}5$ days on 16$\times$40GB A100).
Extension to 7B/14B+ models and to multimodal settings is left as
future work.

\paragraph{Domain coverage.}
Our experiments cover only text-only English mathematical
reasoning across five benchmarks. We do not evaluate STC on code
synthesis, agentic / tool-use tasks, multilingual reasoning,
multimodal (vision+math) problems, or open-ended generation.
While the framework itself is domain-agnostic, our claims about
``critical thinking'' should be read narrowly as referring to
step-level error detection in mathematical chain-of-thought.

\paragraph{Critic reliability.}
STC's step-level balanced
F1 of 67.4\% leaves substantial room for improvement, especially on
Minerva (52.7\% F1) where critic reliability is weakest.

% ===== Backup of original Step granularity paragraph (extended to cover trajectory-style asymmetry per V4.0 reviewer pass) =====
% \paragraph{Step granularity.}
% STC's \texttt{\textbackslash n\textbackslash n}-delimited atomic steps
% (${\sim}250$ per trajectory) are substantially finer than the
% paragraph-level granularity of most external PRMs.
% The PRM comparison (Table~\ref{tab:prm_compare}) therefore operates at
% different base rates across systems and should be read as evidence
% that STC reaches a strong calibration-free operating point, not as a
% direct step-by-step claim over PRMs in their native granularity.
% ===== End of backup =====
\paragraph{Step granularity vs.\ PRMs.}
STC's \texttt{\textbackslash n\textbackslash n}-delimited atomic steps
are substantially finer than the paragraph-level granularity of most
external PRMs.
The PRM comparison (Table~\ref{tab:prm_compare}) therefore operates at
different base rates across systems and should be read as evidence
that STC reaches a strong calibration-free operating point, not as a
direct step-by-step claim over PRMs in their native granularity.

\section{Ethical Considerations}

STC is a methodology study evaluated on standard public mathematical reasoning benchmarks; it does not introduce new user-facing systems, sensitive datasets, or task domains beyond those already established in the LLM reasoning literature. Nevertheless, we identify a concrete risk that warrants disclosure.

\noindent\textbf{Critique reliability.} As shown in Table~\ref{tab:ans_critic_results_wo_VR} of the main manuscript, STC's error-detection accuracy on the answer level remains well below 100\% (averaging 42.6\%); a positive critique should therefore not be interpreted as a correctness guarantee, particularly in high-stakes settings such as education, tutoring, or scientific computing where users may over-trust an explicit score-$1$ signal.

We do not foresee misuse pathways unique to STC beyond those already common to LLM-based mathematical assistants.

% ===============================================================
\bibliography{custom}

% ===============================================================
% \appendix switches numbering to A, B, C...
\appendix
\clearpage
\textbf{\Large Appendix}

\medskip

\noindent\textbf{Overview.} This appendix covers implementation details (\S A), the evaluation protocol (\S B), additional results and qualitative examples (\S C), and discussions on cost and reward hacking (\S D).

% \section{Experiments}  % removed wrapper — subsections promoted to top-level sections

\section{Implementation Details}

\paragraph{SFT data construction.}
\label{sec:sft_data_details}
We randomly select 10k samples and generate one reasoning rollout per sample using the base model. We then filter out samples with reasoning token length larger than 8{,}192, and use GPT-5 to generate critiques for each reasoning step. Next, we apply a three-stage filter to ensure: (i) structural integrity (reasoning and critique blocks are strictly paired); (ii) answer--critique consistency with the ground-truth critic. This yields \textbf{5{,}056} high-quality samples (about 51\% retention). For each retained sample, we interleave reasoning and critique steps to construct a unified trace. The dataset spans a wide range of complexities with token lengths up to 15{,}130: about 75\% of samples fall between 2{,}048 and 8{,}192 tokens, 24\% exceed 8{,}192 tokens, and 10\% exceed 10{,}240 tokens, providing sufficient depth for complex multi-step reasoning.

\paragraph{Hyperparameters and lengths.}
\label{sec:impl_setup}
SFT uses batch size 64 for 550 steps. RL with GRPO uses group size $G{=}8$, batch size 32, $\lambda_\text{format}{=}0.05$, and trains for ${\sim}1200$ steps total (two-stage schedule below). Maximum response lengths are 8{,}192 tokens for reasoning and 4{,}096 tokens for critique. The system prompt is shown in Table~\ref{tab:qualitative} (highlighted in orange).

\paragraph{RL training schedule.}
For RL training, we use a two-stage schedule with different hyperparameter settings. In Stage 1, we set $\lambda_{crit}=0$, $\lambda_{reason}=1$, and $\lambda_{format}=0.05$, and train for about 450 steps to improve reasoning ability and adherence to the reasoning--critique format. In Stage 2, we alternate between two settings every 40 steps for approximately 450 steps: $(\lambda_{reason},\lambda_{format},\lambda_{crit})=(1,0.05,0)$ and $(0,0.05,1)$. This stage is designed to improve both reasoning ability and critique capability.
We alternate these two objectives to mitigate gradient conflict between reasoning and critique optimization.

\paragraph{Full GRPO objective.}
\label{sec:grpo_full}
The GRPO objective maximizes the expected normalized advantage under a KL constraint to the reference policy:
% --- Backup: original single-equation form (overflowed column) ---
% \begin{equation}
% \begin{aligned}
% \mathcal{J}_{\mathrm{GRPO}}(\theta) &=
% \mathbb{E}_{(x,a)\sim \mathcal D,\;\{\tau_k\}_{k=1}^G \sim \pi_{\theta_{\rm old}}(\cdot\mid x)} \\
% &\hspace{-1.5cm}\Bigg[
% \frac1{\sum_{k=1}^G |\tau_k|} \sum_{k=1}^G \sum_{t=1}^{|\tau_k|}
% \bigg(\min\!\Big\{ \rho_{k,t}{\mathcal{A}_{k,t}},\;
% \mathrm{clip}(\rho_{k,t},1-\epsilon,1+\epsilon){\mathcal{A}_{k,t}} \Big\}
% \\
% &\hspace{-1cm} - \beta\,D_{\rm KL}\big( \pi_{\theta}(\tau_k | x) \;\|\; \pi_{\rm ref}(\tau_k | x) \big)\bigg)\Bigg],
% \end{aligned}
% \label{eq:grpo-objective}
% \end{equation}
\begin{equation}
\begin{aligned}
\mathcal{J}_{\mathrm{GRPO}}(\theta) = \mathbb{E}\bigg[
\tfrac{1}{\sum_{k} |\tau_k|}\!\!\sum_{k=1}^G\sum_{t=1}^{|\tau_k|}\!
\Big(\mathcal{L}_{k,t}^{\mathrm{clip}} \\
{}- \beta\, D_{\rm KL}\!\big(\pi_{\theta}(\tau_k|x)\,\|\,\pi_{\rm ref}(\tau_k|x)\big)\Big)\bigg],
\end{aligned}
\label{eq:grpo-objective}
\end{equation}
where the expectation is taken over $(x,a)\sim\mathcal D,\;\{\tau_k\}_{k=1}^G\sim\pi_{\theta_{\rm old}}(\cdot\mid x)$, and $\mathcal{L}_{k,t}^{\mathrm{clip}}=\min\!\big\{\rho_{k,t}\mathcal{A}_{k,t},\;\mathrm{clip}(\rho_{k,t},1-\epsilon,1+\epsilon)\mathcal{A}_{k,t}\big\}$;
$\rho_{k,t}= \frac{\pi_{\theta}({\tau}_{k,t} \mid x, {\tau}_{k,<t})}{{\pi_{\theta_{\text{old}}}}({\tau}_{k,t} \mid x, {\tau}_{k,<t})}$;
$\epsilon$ and $\beta$ are hyperparameters. Here, similar to DAPO~\cite{yu2025dapo}, we use token-level policy gradient loss to optimize the model, where longer sequences can have more influence.

\section{Evaluation Protocol Details}
\label{sec:eval_metrics}
%\subsubsection{Evaluation Protocol} 
We evaluate both the reasoning and critique capability of models. Evaluations are performed with a maximum response length of 8,192 tokens for reasoning and 4,096 tokens for critique by default. 

For problem solving capability, we evaluate model reasoning performance using Pass@1 (P@1 for short) and Pass@8, which capture single-attempt accuracy as well as success rates across multiple sampled attempts.

For critique capability, we evaluate models at both the final answer level and the reasoning-step level. i) For the final answer, the correctness of the critique is \textbf{directly verifiable} (see discussion for Critique-consistency reward in \S\ref{subsub:reward_design}). ii) For intermediate reasoning steps, however, no human-annotated ground truth exists. Therefore, we employ a strong LLM (GPT-5-mini) as an automatic judge to assess the correctness of the model’s stepwise critiques. We denote the pseudo ground-truth critic labels for a trajectory as $z^g = (z_1, z_2, \cdots, z_T)$. Note that we use GPT-5-mini instead of GPT-5 here in considering that GPT-5-mini is much faster than GPT-5 while they achieve similar agreement rate with verifiable critics when evaluated at answer level.

\paragraph{Sampling protocol for critique evaluation.}
To bound LLM-judge cost (GPT-5-mini step labels are billed per token; per-trajectory labeling cost is dominated by the $\sim$250 atomic steps in each trajectory), we evaluate critique on a fixed sample of \textbf{240 STC trajectories per benchmark}: the first 30 prompts of each dataset $\times$ 8 STC rollouts per prompt. This sample is reused for all PRM comparisons (Table~\ref{tab:prm_compare}) and all critic tables (Tables~\ref{tab:ans_critic_results_wo_VR},~\ref{tab:process_critic_results}). AIME24 has exactly 30 prompts and is therefore fully covered; AMC23 (83 prompts), MATH-500 (500), Minerva (272), and OlympiadBench (675) are each represented by their first 30 prompts. Reasoning (Pass@1/Pass@8) is evaluated on each benchmark's full prompt set. Extending critique evaluation to full benchmark coverage is a straightforward follow-up bounded only by LLM-judge cost.  

Since our work is (to our knowledge) the first to unify stepwise reasoning and critique in a single model, there is no off-the-shelf evaluation metric. Previous PRM benchmarks and evaluation metrics \cite{zheng2025processbench} are designed for post-hoc verification and are not suitable for our evaluation since they assume a separate verifier with access to full trajectories and gold step labels, while our setting evaluates a jointly generated step-and-critique process. 

Motivated by \cite{zheng2025processbench}, we design task-aligned metrics that measure correct accuracy (Corr.), error accuracy (Err.), and their harmonic mean (F1). 

\textbf{Correct Accuracy (Corr.).}
Correct accuracy evaluates critic predictions on instances whose ground-truth critic label is 1 (correct).

At the answer level, this measures the fraction of truly correct final answers that are validated as correct by the critic.
Formally, letting $s_{T_j}$ denote the critic’s final prediction for sample $j$ and $z_{T_j}$ its label,
\begin{equation}
\text{Corr.} =
\frac{\sum_{j} \mathbf{1}(s_{T_j} = 1 \land z_{T_j} = 1)}
{\sum_{j} \mathbf{1}(z_{T_j} = 1)}.
\end{equation}
Note that each sample may have a different number of steps $T_j$.

At the process level, this measures the fraction of reasoning steps that are correct according to pseudo ground-truth and are predicted as correct by the critic.
Letting $s_t^{(j)}$ denote the critic’s prediction at step $t$ for sample $j$, we have
\begin{equation}
\text{Corr.} =
\frac{\sum_{j}\sum_{t=1}^{T_j} \mathbf{1}(s_t^{(j)} = 1 \land z_t^{(j)} = 1)}
{\sum_{j}\sum_{t=1}^{T_j} \mathbf{1}(z_t^{(j)} = 1)}.
\end{equation}

\textbf{Error Accuracy (Err.).} Similarly, error accuracy evaluates critic predictions on instances whose ground-truth critic label is 0 (incorrect). 

At the answer level, this measures the fraction of incorrect final answers that are correctly rejected by the critic.
Formally,
\begin{equation}
\text{Err.} =
\frac{\sum_{j} \mathbf{1}(s_{T_j} = 0 \land z_{T_j} = 0)}
{\sum_{j} \mathbf{1}(z_{T_j} = 0)}.
\end{equation}

At the process level, this measures the fraction of reasoning steps that are incorrect according to pseudo ground-truth and are predicted as incorrect by the critic.
Formally,
\begin{equation}
\text{Err.} =
\frac{\sum_{j}\sum_{t=1}^{T_j} \mathbf{1}(s_t^{(j)} = 0 \land z_t^{(j)} = 0)}
{\sum_{j}\sum_{t=1}^{T_j} \mathbf{1}(z_t^{(j)} = 0)}.
\end{equation}

\textbf{F1.}
Since critique evaluation is inherently imbalanced (correct instances often dominate), we adopt the harmonic mean of correct accuracy and error accuracy for overall critic capability evaluation:
\[
\text{Balanced F1} = \frac{2 \cdot \text{Corr.} \cdot \text{Err.}}
{\text{Corr.} + \text{Err.}}.
\]
This metric penalizes degenerate solutions that achieve high correct accuracy but fail to detect errors, or vice versa.

\section{Additional Results}

\subsection{Reliability of GPT-5-mini as a Judge}
\label{sec:claude_reliability}

\begin{table}[t]
\centering
\scriptsize
\setlength{\tabcolsep}{4pt}
\renewcommand{\arraystretch}{1.05}
\begin{tabular*}{\columnwidth}{l@{\extracolsep{\fill}}*{6}{r}}
\toprule
& \multicolumn{3}{c}{\textbf{GPT-5-mini reference}} & \multicolumn{3}{c}{\textbf{Claude reference}} \\
\cmidrule(lr){2-4}\cmidrule(lr){5-7}
\textbf{Dataset} & Corr.\ & Err.\ & F1 & Corr.\ & Err.\ & F1 \\
\midrule
\textbf{AIME24}      & 84.4 & 61.3 & 71.0 & 80.1 & 64.4 & \textbf{71.4} \\
\bottomrule
\end{tabular*}
\caption{\textbf{STC process-level critic performance under each LLM reference (AIME24).} All values in \%. The same STC predictions are scored against either GPT-5-mini or Claude Opus 4.7 step labels. Switching the reference changes Corr.\ by $-4.3\%$ and Err.\ by $+3.1\%$, with balanced F1 shifting by $0.4\%$ on AIME24.}
\label{tab:stc_under_two_judges}
\end{table}

% ===== Tab 6 source moved here (was at start of \S\ref{sec:more_stats}) so it enters the float queue earlier and lands at p.13 top alone, before Tab 7 starts competing =====
\begin{table*}[!t]
\centering
\scriptsize
\setlength{\tabcolsep}{4pt}
\renewcommand{\arraystretch}{1.05}
\begin{tabular*}{\linewidth}{ll@{\extracolsep{\fill}}rrrrccc}
\toprule
\textbf{Method} & \textbf{Dataset}
& \textbf{TP} & \textbf{FP} & \textbf{TN} & \textbf{FN}
& \textbf{Valid Ratio}
& \textbf{GT Neg. Ratio}
& \textbf{Pred. Neg. Ratio} \\
\midrule

\multirow{6}{*}{STC-SFT}
 & AIME24 & 44 & 17 & 2 & 0 & 26.3 & 27.0 & 3.2 \\
 & AMC23 & 321 & 24 & 7 & 5 & 53.8 & 8.1 & 3.4 \\
 & MATH & 2907 & 224 & 18 & 25 & 79.4 & 7.8 & 1.4 \\
 & Minerva & 425 & 777 & 74 & 15 & 59.3 & 61.4 & 6.9 \\
 & OlympiadBench & 1820 & 754 & 64 & 28 & 49.4 & 29.3 & 3.5 \\
 & \textbf{Average} & -- & -- & -- & -- & 53.6 & 27.0 & 4.0 \\
\midrule

\multirow{6}{*}{STC w/o Critic}
 & AIME24 & 66 & 52 & 18 & 2 & 57.5 & 39.1 & 14.5 \\
 & AMC23 & 396 & 88 & 32 & 20 & 80.7 & 20.1 & 9.7 \\
 & MATH & 3223 & 377 & 75 & 87 & 94.1 & 12.3 & 4.3 \\
 & Minerva & 558 & 1223 & 178 & 32 & 91.5 & 63.0 & 10.5 \\
 & OlympiadBench & 2190 & 1535 & 490 & 134 & 80.5 & 38.4 & 14.3 \\
 & \textbf{Average} & -- & -- & -- & -- & 80.9 & 35.0 & 11.0 \\
\midrule

\multirow{6}{*}{STC}
 & AIME24 & 55 & 23 & 35 & 12 & 52.1 & 28.0 & 37.6 \\
 & AMC23 & 363 & 48 & 59 & 39 & 76.7 & 17.1 & 19.3 \\
 & MATH & 3188 & 308 & 149 & 127 & 94.3 & 11.5 & 7.3 \\
 & Minerva & 545 & 1053 & 295 & 60 & 89.8 & 57.0 & 18.2 \\
 & OlympiadBench & 2084 & 1028 & 777 & 218 & 76.1 & 30.3 & 24.2 \\
 & \textbf{Average} & -- & -- & -- & -- & 77.8 & 29.0 & 21.0 \\
% \midrule

\bottomrule
\end{tabular*}
\caption{\textbf{Answer-level confusion and statistics.} All ratios in \%. Valid Ratio: fraction of samples where the final critic could be extracted (failures usually due to token overlength or unmatched \texttt{<score></score>}); we only count valid samples. \emph{GT Neg. Ratio} is the ground-truth negative ($z{=}0$) rate; \emph{Pred. Neg. Ratio} is the predicted negative rate. The base model DS-Qwen-1.5B has 78.4\% average valid ratio (due to over-length).}
\label{tab:answer_level_confusion}
\end{table*}

% Tab 7 source also moved into \S\ref{sec:claude_reliability} so Tab 6 + Tab 7 enter the queue together and both land on p.13 top.
\begin{table*}[!t]
\centering
\scriptsize
\setlength{\tabcolsep}{4pt}
\renewcommand{\arraystretch}{1.05}
\begin{tabular*}{\linewidth}{ll@{\extracolsep{\fill}}rrrrcc}
\toprule
\textbf{Method} & \textbf{Dataset}
& \textbf{TP} & \textbf{FP} & \textbf{TN} & \textbf{FN}
& \textbf{GT Neg. Ratio} & \textbf{Pred. Neg. Ratio} \\
\midrule

\multirow{6}{*}{SFT}
 & AIME24 & 38654 & 7471 & 4102 & 2644 & 21.9 & 12.8 \\
 & AMC23 & 31686 & 4487 & 2049 & 1862 & 16.3 & 9.8 \\
 & MATH & 19737 & 2228 & 713 & 559 & 12.7 & 5.5 \\
 & Minerva & 29822 & 8242 & 1976 & 1455 & 24.6 & 8.3 \\
 & OlympiadBench & 32043 & 6818 & 5567 & 3322 & 25.9 & 18.6 \\
 & \textbf{Average} & 30388 & 5849 & 2881 & 1968 & 20.3 & 11.0 \\
\midrule

\multirow{6}{*}{STC w/o Critic}
 & AIME24 & 40348 & 6438 & 3496 & 2568 & 18.8 & 11.5 \\
 & AMC23 & 28724 & 4469 & 1918 & 1577 & 17.4 & 9.5 \\
 & MATH & 17185 & 1656 & 552 & 694 & 11.0 & 6.2 \\
 & Minerva & 24808 & 6705 & 1715 & 1424 & 24.3 & 9.1 \\
 & OlympiadBench & 32281 & 5956 & 5335 & 3628 & 23.9 & 19.0 \\
 & \textbf{Average} & 28669 & 5045 & 2603 & 1978 & 19.1 & 11.1 \\
\midrule

\multirow{6}{*}{STC}
 & AIME24 & 41205 & 4471 & 7082 & 7612 & 19.1 & 24.3 \\
 & AMC23 & 29572 & 3098 & 4558 & 5652 & 17.9 & 23.8 \\
 & MATH & 15815 & 1178 & 1343 & 1809 & 12.5 & 15.7 \\
 & Minerva & 25370 & 4661 & 2772 & 2948 & 20.8 & 16.0 \\
 & OlympiadBench & 32718 & 3496 & 10634 & 9261 & 25.2 & 35.5 \\
 & \textbf{Average} & 28936 & 3381 & 5278 & 5456 & 19.1 & 23.1 \\
% \midrule

% \multirow{6}{*}{STC-GRPO}
%  & AIME24 & 29611 & 1976 & 8561 & 18457 & 18.0 & 46.1 \\
%  & AMC23 & 23626 & 1712 & 5158 & 10849 & 16.6 & 38.7 \\
%  & MATH & 14165 & 725 & 1721 & 3863 & 12.0 & 27.3 \\
%  & Minerva & 20562 & 3188 & 3943 & 6287 & 21.0 & 30.1 \\
%  & OlympiadBench & 23742 & 1448 & 10435 & 20909 & 21.0 & 55.4 \\
%  & \textbf{Average} & 22341 & 1810 & 5964 & 12073 & 17.7 & 39.5 \\
\bottomrule
\end{tabular*}
\caption{\textbf{Process-level confusion and statistics.} Computed on the first 240 trajectories (30 prompts $\times$ 8 rollouts) per benchmark.}
\label{tab:process_level_confusion}
\end{table*}

% Fig 4 source moved here (from \S\ref{sec:prm_answer_level}) so it enters the float queue early and lands at p.14 top.
\begin{figure*}[!t]
\centering
\includegraphics[width=\linewidth]{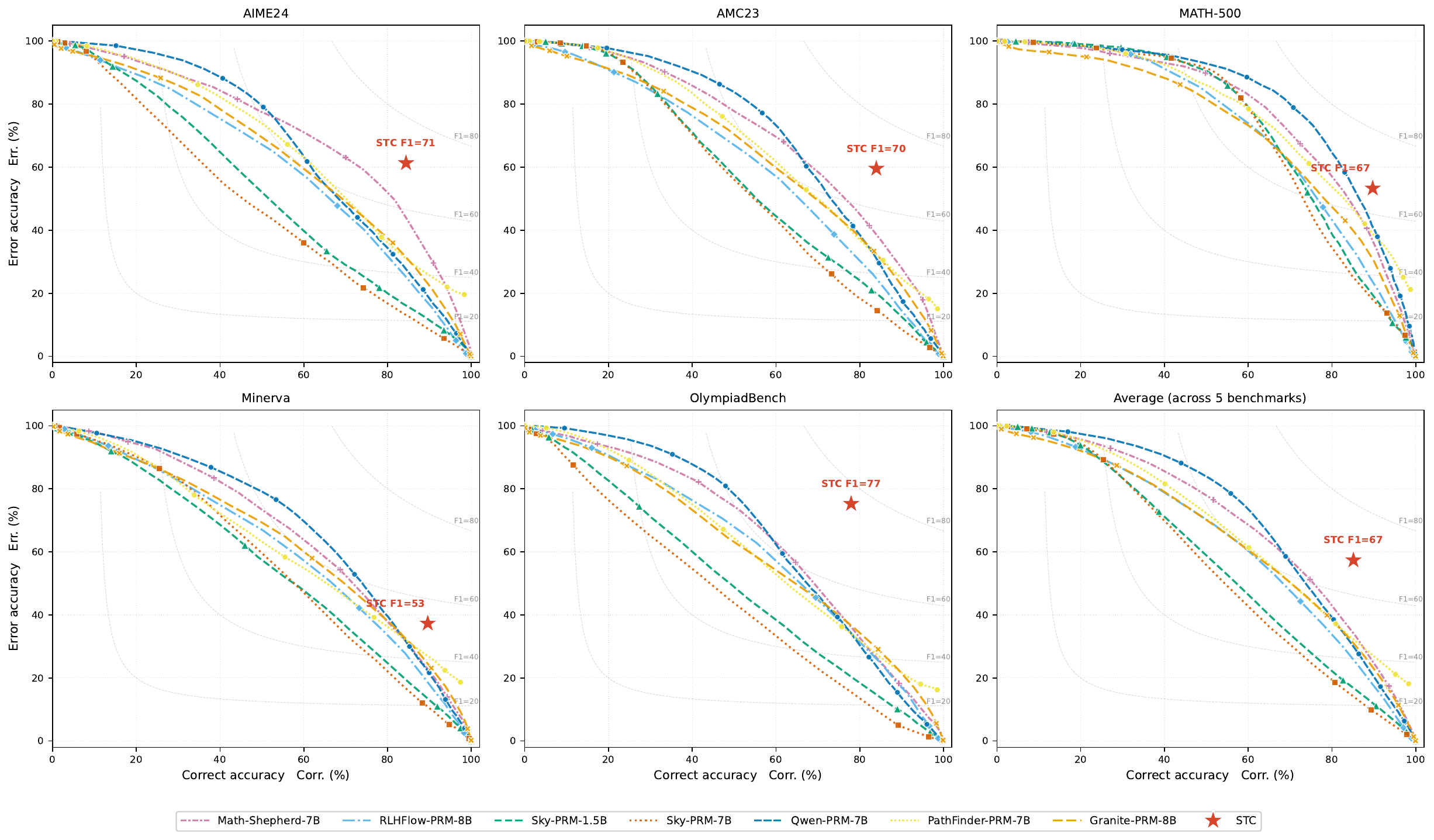}
\caption{\textbf{Step-level Corr.\ vs.\ Err.\ trade-off across 5 benchmarks.} STC (\textcolor{red}{$\bigstar$}, single calibration-free point with annotated F1) vs.\ seven external PRMs (curves swept over $\sim$200 thresholds in $[0.001, 0.999]$). The \emph{Average} subplot is the cross-dataset arithmetic mean. \emph{Legend ``Qwen-PRM-7B'' refers to Qwen2.5-Math-PRM-7B.}}
\label{fig:prm_corr_err_curves_step}
\end{figure*}

In the \emph{Reliability of the GPT-5-mini judge} paragraph (\S\ref{subsub:critique}) of the main text, we assess the reliability of GPT-5-mini step labels by comparing with verifiable answer-level ground truth (about 90\% agreement). Since this is an indirect, answer-level check, we further perform a \emph{direct step-level} validation by re-evaluating the AIME24 labeled corpus with a stronger LLM judge --- \textbf{Claude Opus 4.7} (with maximum reasoning budget and no output truncation) --- under the same per-step prompt as GPT-5-mini (\textbf{60k+} step labels in total).

\textbf{GPT-5-mini achieves balanced F1 of 87.3\% (Corr.\ 88.3\%, Err.\ 86.4\%, raw agreement 88.1\%) against Claude as gold standard on AIME24.} The disagreement is asymmetric: GPT-5-mini calls $19.2\%$ of steps wrong whereas Claude calls only $10.1\%$, \ieno, GPT-5-mini is a more conservative judge that over-flags errors by roughly $1.9\times$, but it almost never \emph{misses} a Claude-flagged error.

\paragraph{STC critic performance under two judges.}
Table~\ref{tab:stc_under_two_judges} re-computes STC's process-level critic performance on AIME24 using \emph{either} GPT-5-mini or Claude as the reference. On AIME24, balanced F1 differs by only $0.4\%$ ($71.0\% \to 71.4\%$). The two metrics move in opposite directions when the reference is swapped: Corr.\ drops $4.3\%$ (Claude's larger correct-class makes a stricter denominator) while Err.\ rises $3.1\%$ (Claude's wrong-class is smaller and cleaner).

% ===== Original Tab 6 + Tab 7 position (moved to start of \S\ref{sec:more_stats} for p.13 top placement) =====

\subsection{More Statistics for Critiques}
\label{sec:more_stats}

% Tab 6 and Tab 7 source moved to \S\ref{sec:claude_reliability} for p.13 placement.

Table~\ref{tab:answer_level_confusion} and Table~\ref{tab:process_level_confusion} show the answer-level and step-level confusion matrixes and statistics.

In Table~\ref{tab:answer_level_confusion}, to ensure transparency, we additionally report the valid ratio (VR), \ieno, the proportion of samples feasible for evaluation of the critique performance. In general, the failures in extracting final critic are treated as invalid samples, which are usually due to token overlength or unmatched \texttt{<score></score>}.  
Different schemes present different valid ratios. Since there are large differences on the valid ratios between \emph{STC-SFT} and the other two schemes, the two schemes \emph{STC w/o Critic} and \emph{STC} with similar valid ratios are more fairly comparable to analyse the performance.

Since critique performance is computed only on valid samples, VR contextualizes comparability across methods. 

For step-level evaluation, since each critique is following a reasoning step, the reasoning-critique pairs even for over-length traces are also used for evaluation. Therefore we do not calculate the valid ratio.

\subsection{More Comparisons of STC and External PRMs}
\label{sec:prm_answer_level}

We complement Table~\ref{tab:prm_compare} with two additional comparison axes: (i) full Corr.--Err.\ trade-off curves under threshold sweep, and (ii) inference compute and memory.

\begin{figure*}[!t]
\centering
\includegraphics[width=\linewidth]{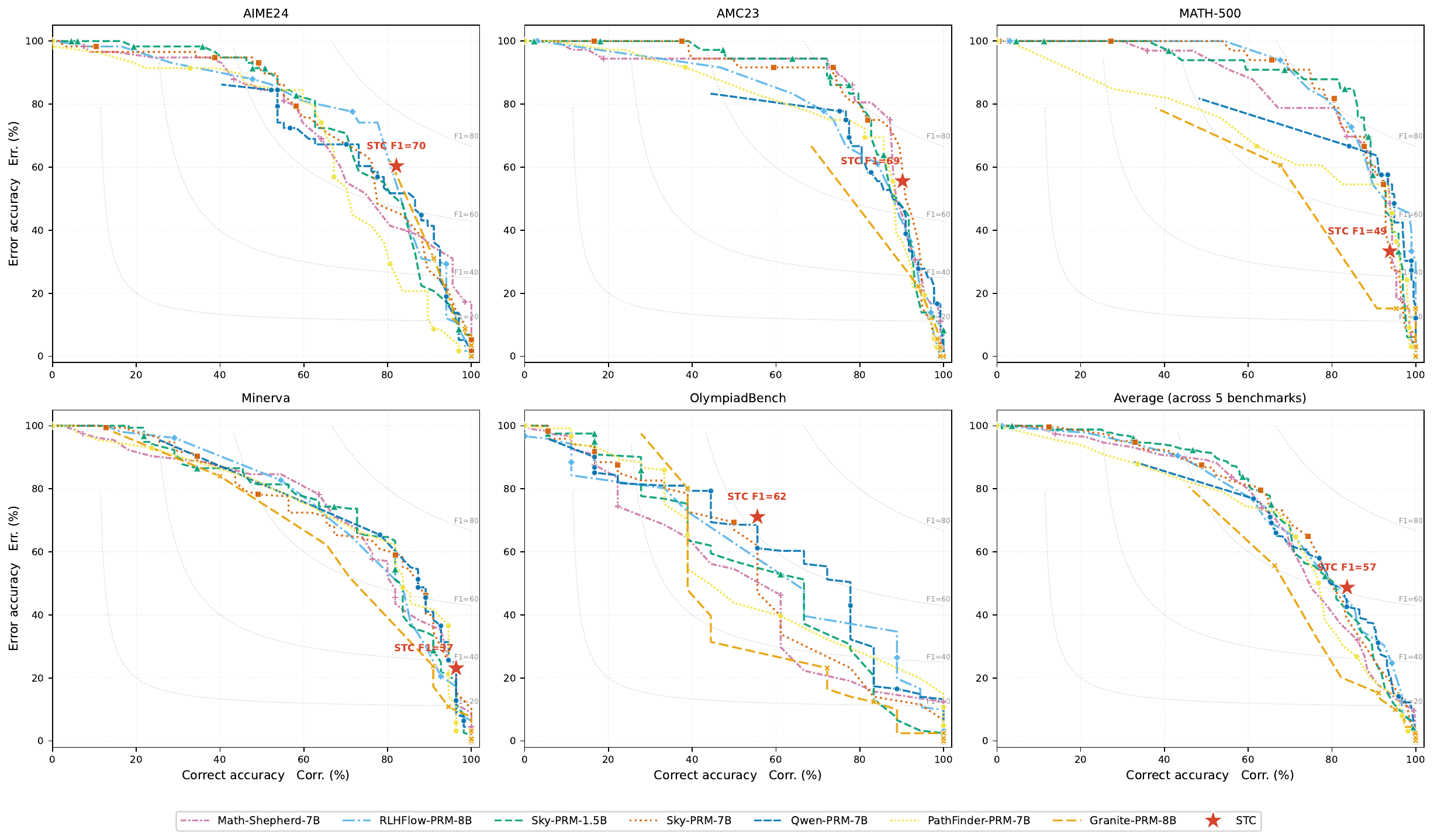}
\caption{\textbf{Answer-level Corr.\ vs.\ Err.\ trade-off across 5 benchmarks.} Same protocol as Figure~\ref{fig:prm_corr_err_curves_step}.}
\label{fig:prm_corr_err_curves_answer}
\end{figure*}

\paragraph{Accuracy: Corr.\ vs.\ Err.\ trade-off curves.}
Figures~\ref{fig:prm_corr_err_curves_step} and~\ref{fig:prm_corr_err_curves_answer} show each PRM's full Corr.--Err.\ trade-off under threshold sweep (on 30 prompts $\times$ 8 rollouts $=$ 240 trajectories), at the step level and answer level respectively. The seven PRMs span training context lengths of 4K--128K; positions beyond the trained context are handled via RoPE extrapolation.

% Fig 4 source moved earlier to \S\ref{sec:claude_reliability}; Fig 5 stays here.

% Tab 8 source moved here (right after Fig 5) so it queues right after Fig 5; with [!t] this places it at the right-column top of p.14 (below Fig 4) -- on the same page as the in-text reference.
\begin{table}[!t]
\centering
\scriptsize
\setlength{\tabcolsep}{4pt}
\renewcommand{\arraystretch}{1.05}
\begin{tabular*}{\columnwidth}{l@{\extracolsep{\fill}}rrrcc}
\toprule
\multirow{2}{*}{\textbf{Setup}} & \textbf{Params} & \textbf{FLOPs} & \textbf{Mem} & \multicolumn{2}{c}{\textbf{STC saves}} \\
\cmidrule(lr){5-6}
 & (B) & (T) & (GiB) & FLOPs & Mem \\
\midrule
STC (single 1.5B)         & 1.5 & 26.0  & 3.0  & ---  & ---  \\
\midrule
1.5B + Skywork-PRM-1.5B   & 3.0 & 36.1  & 5.9  & 28\% & 49\% \\
1.5B + 7B PRM             & 8.5 & 102.3 & 16.3 & 75\% & 82\% \\
1.5B + 8B PRM             & 9.5 & 114.3 & 18.6 & 77\% & 84\% \\
\bottomrule
\end{tabular*}
\caption{\textbf{Per-trajectory inference cost: STC vs separate-verifier PRM pipelines.} FLOPs computed as $2N_\text{params}T$ with measured $R{=}6{,}016$ reasoning tokens and $C{=}2{,}638$ critique tokens on AIME24. Peak memory is fp16 weights plus GQA KV cache\tablefootnote{The KV cache is $<$8\% of total memory in every configuration because all baselines use grouped-query attention (e.g., 2 KV heads for the 1.5B model), so peak memory is dominated by model weights.} for an 8K-token trajectory at batch size 1, assuming concurrent serving (both reasoner and PRM resident on the GPU). 7B PRMs: Math-Shepherd, RLHFlow, Qwen2.5-Math-PRM, PathFinder. 8B PRM: Granite-3.3.}
\label{tab:inference_cost}
\end{table}

% Tab 9 source moved here (between \S\ref{sec:prm_answer_level} end and \S\ref{sec:pass_at_k_variability} start) so it enters the queue late enough to skip p.14 (occupied by Tab 8) and land at the right-column top of p.15 below Fig 5.
% Removed previous Tab 9 block (was right after Tab 8) to avoid duplicate placement.

\paragraph{Step level.} STC's single calibration-free point sits on or above every PRM's full sweep on AIME24, AMC23, and OlympiadBench, and is competitive on MATH-500 and Minerva. In the \emph{Average} panel, STC reaches the best step-F1 in expectation.

\paragraph{Answer level.} STC's point lies on the PRM frontier on most datasets (four out of five), aligned with the best-performing PRM curves. The answer-level curves are visibly less smooth than the step-level ones because each curve is now built from only 240 trajectory outcomes per benchmark, versus tens of thousands of step labels at the step level (Table~\ref{tab:process_level_confusion}).

\paragraph{Inference compute and memory.}
A separate-verifier pipeline must load and run two models: a reasoner ($N_R$ params) generates $R$ reasoning tokens autoregressively, then a PRM ($N_{PRM}$ params) prefills the trajectory to score each step, costing approximately $2R(N_R{+}N_{PRM})$ FLOPs. STC instead emits $R$ reasoning tokens and $C$ critique tokens inline within a single 1.5B model, costing $\approx 2N_R(R{+}C)$ FLOPs. Empirically, on the 240 STC trajectories from AIME24 (30 prompts $\times$ 8 rollouts), $C/R \approx 0.44$. Table~\ref{tab:inference_cost} summarizes the resulting per-trajectory cost: STC saves $\sim$28\% FLOPs and $\sim$49\% peak memory against the smallest 1.5B-PRM baseline, and $\sim$75--77\% FLOPs and $\sim$82--84\% peak memory against the 7B/8B baselines. STC additionally avoids the system overhead of dispatching trajectories between two models (scheduling and cross-model serialization).

% Tab 8 source moved earlier (right after Fig 5) so it lands at the right-column top of p.14 (below Fig 4), next to its in-text reference.

\begin{table}[!t]
\centering
\scriptsize
\setlength{\tabcolsep}{4pt}
\renewcommand{\arraystretch}{1.05}
\begin{tabular*}{\columnwidth}{l@{\extracolsep{\fill}}cccc}
\toprule
\multirow{2}{*}{\textbf{Eval setup}}
& \multicolumn{2}{c}{\textbf{AIME24}} & \multicolumn{2}{c}{\textbf{AMC23}} \\
\cmidrule(lr){2-3}\cmidrule(lr){4-5}
& P@1 & P@8 & P@1 & P@8 \\
\midrule
1 seed & 27.9 & 56.7 & 60.5 & 81.9 \\
\addlinespace[1pt]
16 seeds (mean$\pm$std) & 27.7\,$\pm$\,2.0 & 56.5\,$\pm$\,5.5 & 62.3\,$\pm$\,0.9 & 83.7\,$\pm$\,1.8 \\
\bottomrule
\end{tabular*}
\caption{\textbf{Pass@1 / Pass@8 (\%) variability of STC (full) on AIME24 and AMC23, measured by mean$\pm$std across 16 disjoint 8-rollout seeds ($N{=}128$ rollouts/prompt).} The 1-seed row reproduces values from Table~\ref{tab:main-acc}; the 16-seed row uses the same checkpoint.}
\label{tab:pass_at_k_variability}
\end{table}

% Tab 10 source moved here (right after Tab 9, in \S\ref{sec:prm_answer_level}) so it queues right after Tab 9 and lands at p.16 top.
\begin{table*}[!t]
\centering
\scriptsize
\setlength{\tabcolsep}{4pt}
\renewcommand{\arraystretch}{1.05}
\begin{tabular*}{\linewidth}{l@{\extracolsep{\fill}}*{6}{cc}}
\toprule
\multirow{2}{*}{\textbf{Model}}
& \multicolumn{2}{c}{\textbf{AIME24}} & \multicolumn{2}{c}{\textbf{AMC23}}
& \multicolumn{2}{c}{\textbf{MATH-500}} & \multicolumn{2}{c}{\textbf{Minerva}}
& \multicolumn{2}{c}{\textbf{OlympiadBench}} & \multicolumn{2}{c}{\textbf{Average}} \\
\cmidrule(lr){2-3}\cmidrule(lr){4-5}\cmidrule(lr){6-7}\cmidrule(lr){8-9}\cmidrule(lr){10-11}\cmidrule(lr){12-13}
& P@1 & P@8 & P@1 & P@8 & P@1 & P@8 & P@1 & P@8 & P@1 & P@8 & P@1 & P@8 \\
\midrule
STC          & 27.9          & \textbf{56.7} & 60.5          & 81.9          & \textbf{82.9} & \textbf{92.6} & \textbf{27.9} & \textbf{49.3} & 42.7          & 57.3          & 48.4          & \textbf{67.6} \\
\addlinespace[2pt]
STC w/ Dense & \textbf{30.4} & 53.3          & \textbf{64.2} & \textbf{83.1} & 82.6          & 92.0          & 27.3          & 48.2          & \textbf{43.6} & \textbf{59.0} & \textbf{49.6} & 67.1          \\
\bottomrule
\end{tabular*}
\caption{\textbf{Influence of stepwise critique as dense reward.} Reasoning performance (\%) on five math benchmarks, evaluated with Pass@1 (P@1) and Pass@8 (P@8). \textbf{Bold} marks the better of the two per column.}
\label{tab:dense_reward}
\end{table*}

\subsection{Pass@$k$ Variability Across Evaluation Seeds}
\label{sec:pass_at_k_variability}

AIME24 (30 problems) and AMC23 (83 problems) are small benchmarks of the five testing datasets; a single $n{=}8$ rollout evaluation can be noisy. In contrast, MATH-500, Minerva, and OlympiadBench contain 500, 272, and 675 problems respectively, so their single-seed estimates are much more robust. To quantify the small-benchmark noise, we extend the rollout budget to $n{=}128$ per prompt for the STC (full) checkpoint on AIME24 and AMC23, and partition the 128 rollouts into 16 disjoint 8-rollout splits, treated as 16 independent eval seeds. Table~\ref{tab:pass_at_k_variability} compares the original single-seed result (the value reported in Table~\ref{tab:main-acc}) against the 16-seed mean$\pm$std, where std is the spread of a single 8-rollout evaluation across the 16 seeds. The std quantifies how far an arbitrary single-seed evaluation may drift.

On AIME24, the 16-seed mean almost coincides with the single-seed value (shifts within $\sim$0.2\,\%, negligible against the single-seed std of $\sim$2--6\,\%). On AMC23, the 16-seed mean is about 1.8\,\% higher than the single seed results reported in Table~\ref{tab:main-acc}.

% Tab 9 and Tab 10 source moved earlier (right after each other in \S\ref{sec:prm_answer_level}) for p.15 / p.16 top placement.

\subsection{Discussion on Stepwise Critique as Dense Rewards}
\label{sec:dense_reward_discussion}

% Tab 10 source moved earlier (into \S\ref{sec:pass_at_k_variability}) for p.16 top placement.

Alternatively, we leverage stepwise critique judgments as dense rewards to further improve learning of the reasoning process.

%\noindent\emph{\textbf{stepwise critique as dense rewards.}} 
%While the trajectory-level critique-consistency reward provides global supervision, it may be too sparse to guide intermediate reasoning steps effectively.  
%We further leverage the stepwise judgements of the critic as dense rewards to promote the  reasoning process learning.  

Concretely, the critique model produces a binary decision 
$s_{n}^{(k)} \in \{0,1\}$ at each reasoning step $n$ of trajectory $\tau_k$, 
indicating whether the reasoning step is correct or not. 
We obtain step-level advantages by summing the normalized step rewards of its following steps:
\begin{equation}
A_{k}^{dense,n} = \sum_{n' \geq n}
\frac{{s_{n'}^{(k)}}  - \mathrm{mean}\!\left(\mathcal{S}\right)}
     {\mathrm{std}\!\left(\mathcal{S}\right)}, 
\end{equation}
where $\mathcal{S} = \{{s_{n}^{(k)}}|n = 1,2,\cdots, T^{(k)}, k=1,2,\cdots,G\}$ denotes the set including all step critique rewards within the group. 

The stepwise reasoning advantage for the $t$-th token is 
\begin{equation}
    \mathcal{A}^{\text{dense}}_{k,t} = A_{k}^{\text{dense},s(t)},
\end{equation} 
where $s(t)$ denotes the identification of the reasoning step index that the $t$-th token belongs to.
%; if it does not belong to reasoning steps (e.g., belongs to critique steps), $\mathcal{A}^{\text{step}}_{k,t}=0$. 

The overall advantage for the $t$-th token of the trajectory $\tau_k$ is then augmented by this dense critique signal as:  
\begin{equation}
    \widetilde{\mathcal{A}_{k,t}} = \mathcal{A}_{k,t} + \lambda_{\text{dense}}\mathcal{A}^{\text{dense}}_{k,t},
\end{equation}
where $\lambda_{\text{dense}}$ balances the contribution of dense stepwise feedback ($\lambda_\text{dense} = 0.5$). 

This extension transforms the optimization from relying solely on sparse trajectory-level rewards to incorporating fine-grained, step-level supervision.  
By providing dense reward shaping, the model receives immediate reward signals for fine-grained optimization.

Finally, the GRPO objective maximizes the expected normalized advantage under a KL constraint to the reference policy:
% --- backup of original (18); formula was too wide ---
% \begin{equation}
% \begin{aligned}
% \mathcal{J}_{\mathrm{GRPO}}(\theta) &=
% \mathbb{E}_{(x,a)\sim \mathcal D,\;\{\tau_k\}_{k=1}^G \sim \pi_{\theta_{\rm old}}(\cdot\mid x)} \\
% &\hspace{-1.5cm}\Bigg[
% \frac1{\sum_{k=1}^G |\tau_k|} \sum_{k=1}^G \sum_{t=1}^{|\tau_k|}
% \bigg(\min\!\Big\{ \rho_{k,t}\widetilde{\mathcal{A}_{k,t}},\;
% \mathrm{clip}(\rho_{k,t},1-\epsilon,1+\epsilon)\widetilde{\mathcal{A}_{k,t}} \Big\}
% \\
% &\hspace{-1cm} - \beta\,D_{\rm KL}\big( \pi_{\theta}(\tau_k | x) \;\|\; \pi_{\rm ref}(\tau_k | x) \big)\bigg)\Bigg],
% \end{aligned}
% \label{eq:grpo-objective-prm}
% \end{equation}
% where $\rho_{k,t}= \frac{\pi_{\theta}({\tau}_{k,t} \mid x, {\tau}_{k,<t})}{{\pi_{\theta_{\text{old}}}}({\tau}_{k,t} \mid x, {\tau}_{k,<t})}$.
% $\epsilon$ and $\beta$ are hyperparameters.
\begin{equation}
\begin{aligned}
\mathcal{J}_{\mathrm{GRPO}}(\theta) = \mathbb{E}\bigg[
\tfrac{1}{\sum_{k} |\tau_k|}\!\!\sum_{k=1}^G\sum_{t=1}^{|\tau_k|}\!
\Big(\widetilde{\mathcal{L}}_{k,t}^{\mathrm{clip}} \\
{}- \beta\, D_{\rm KL}\!\big(\pi_{\theta}(\tau_k|x)\,\|\,\pi_{\rm ref}(\tau_k|x)\big)\Big)\bigg],
\end{aligned}
\label{eq:grpo-objective-prm}
\end{equation}
where the expectation is taken over $(x,a)\sim\mathcal D,\;\{\tau_k\}_{k=1}^G\sim\pi_{\theta_{\rm old}}(\cdot\mid x)$, and $\widetilde{\mathcal{L}}_{k,t}^{\mathrm{clip}}=\min\!\big\{\rho_{k,t}\widetilde{\mathcal{A}_{k,t}},\;\mathrm{clip}(\rho_{k,t},1-\epsilon,1+\epsilon)\widetilde{\mathcal{A}_{k,t}}\big\}$;
$\rho_{k,t}= \frac{\pi_{\theta}({\tau}_{k,t} \mid x, {\tau}_{k,<t})}{{\pi_{\theta_{\text{old}}}}({\tau}_{k,t} \mid x, {\tau}_{k,<t})}$;
$\epsilon$ and $\beta$ are hyperparameters.

% ===== Original Table 9 position (moved up to top of \S\ref{sec:dense_reward_discussion} for p.16 top placement) =====

We implemented this and Table~\ref{tab:dense_reward} shows the results. Compared with plain STC, dense reward supervision shifts average Pass@1 by $+1.2\%$ and average Pass@8 by $-0.5\%$.

We thus thought our existing implementation does not achieve a robust gain. More effective reward shaping may be needed to unlock the potential of stepwise critique as a dense reward, which we leave as future work.

% ===== OLD tab:example-compact (dense-reward STC compact mode) commented out; replaced by 4 plain-STC tables below =====
\subsection{More Qualitative Examples}
\label{sec:more_qualitative}

Tables~\ref{tab:qualitative} to~\ref{tab:example_olympiad_primes} present four full-mode trajectories spanning four benchmarks (AIME24, Minerva, MATH-500, OlympiadBench), illustrating the qualitative behaviors that STC's critic exhibits in practice.

Across these examples the critique is concise and informative: it pinpoints the \emph{cause} of a detected error rather than emitting a binary flag. The four examples illustrate, but are not limited to, the following error types that STC successfully identifies:
(i) \textbf{Invalid mathematical shortcut} -- Table~\ref{tab:qualitative} (AIME24, lottery probability): STC rejects a bogus alternative formula $1/\binom{6}{2}$ that the model briefly floats, pointing back at the prior ``given at least two'' enumeration; the trajectory re-derives the correct answer $1/115$.
(ii) \textbf{Meta-reasoning over irrelevant context} -- Table~\ref{tab:example_minerva_dish} (Minerva): STC identifies that the interferometer description is irrelevant to a \emph{single}-dish-resolution question.
(iii) \textbf{Constraint misreading} -- Table~\ref{tab:example_math_army} (MATH-500, Greek army combinations): STC catches that the army must field \emph{exactly} the demanded force, not ``meet or exceed'' it.
(iv) \textbf{Factual and symbolic errors in long trajectories} -- Table~\ref{tab:example_olympiad_primes} (OlympiadBench, prime $y_n$ sequence, 188 steps): STC catches two independent errors, a wrong uniqueness claim about $a{=}1$ and a wrong expression for $y_2$.
Color conventions: critiques identified as correct steps are shown in \textcolor{criticcolor}{green}, critiques identified as incorrect steps in \textcolor{criticcolor_0}{red}, and the system prompt in \textcolor{orange}{orange}. For readability we display the first and last few steps verbatim, plus a $\pm 1$-step window around each \scorezero, and skip the (uniformly correct) intermediate steps.

\begin{table*}[!t]
\centering
\tiny
\begin{tabularx}{1\textwidth}{>{\raggedright\arraybackslash}X}
\toprule
\textbf{Stepwise Think-Critique (Full Mode) -- AIME24 example} \\
\midrule
\addlinespace

<|User|>: Jen enters a lottery by picking $4$ distinct numbers from $S=\{1,2,3,\cdots,9,10\}.$ $4$ numbers are randomly chosen from $S.$ She wins a prize if at least two of her numbers were $2$ of the randomly chosen numbers, and wins the grand prize if all four of her numbers were the randomly chosen numbers. The probability of her winning the grand prize given that she won a prize i... \\
\addlinespace
\textcolor{orange}{Let's think step by step and output the final answer within \textbackslash \textbackslash boxed\{\}.}

\textcolor{orange}{Leverage the output \texttt{\textbackslash n\textbackslash n} as the indicator for separating reasoning steps. Following each reasoning step, please provide a critique and a score, where the critique reasons about the correctness of this step within \texttt{<critic>} and \texttt{</critic>} tags, the score of 1 for denoting correct or 0 denoting incorrect within \texttt{<score>} and \texttt{</score>} tags. Follow the format in this example precisely: [Your thinking process for a step goes here] \texttt{<critic>} [Your brief justification for the score goes here] \texttt{</critic>} \texttt{<score>} [0 or 1] \texttt{</score>}} \\

\arrayrulecolor{gray}\midrule\arrayrulecolor{black}
\addlinespace

<|Assistant|><think> \\
\addlinespace
\textcolor{gray}{Step 1.} Alright, so I've got this probability problem here, and I need to figure it out step by step. Let me try to understand what's being asked. \\
\textcolor{criticcolor}{<critic>Clear setup and understanding of the problem context.</critic>} \scoreone \textcolor{gray}{[GPT-5-mini: 1]} \\
\addlinespace
\textcolor{gray}{Step 2.} Jen is entering a lottery by picking 4 distinct numbers from the set S = {1, 2, 3, ..., 9, 10}. That means S has 10 numbers, right? So, the total number of possible ways she can pick her 4 numbers is the combination of 10 taken 4 at a time. I remember that the formula for combinations is C(n, k)... \\
\textcolor{criticcolor}{<critic>Correct identification of total combinations C(10,4).</critic>} \scoreone \textcolor{gray}{[GPT-5-mini: 1]} \\
\addlinespace
\textit{(\dots skipping 27 thinking steps \dots)} \\
\addlinespace
\textcolor{gray}{Step 30.} Yes, that seems correct. \\
\textcolor{criticcolor}{<critic>Affirmation of correctness; consistent with prior results.</critic>} \scoreone \textcolor{gray}{[GPT-5-mini: 1]} \\
\addlinespace
\textcolor{gray}{Step 31.} Alternatively, another way to think about it is that given that at least two numbers are correct, the probability that all four are correct is 1 / C(6, 2). Wait, no, that might not be the right way. \\
\textcolor{criticcolor_0}{<critic>The suggested alternative (1/C(6,2)) is not valid; the ``given at least two'' reasoning is fine.</critic>} \scorezero \textcolor{gray}{[GPT-5-mini: 0]} \\
\addlinespace
\textcolor{gray}{Step 32.} Wait, perhaps I can compute it as the ratio of the number of ways to have exactly 4 correct over the number of ways to have at least 2 correct. \\
\textcolor{criticcolor}{<critic>Correct ratio interpretation.</critic>} \scoreone \textcolor{gray}{[GPT-5-mini: 1]} \\
\addlinespace
\textcolor{gray}{Step 33.} So, the number of ways to have exactly 4 correct is 1, as before. The number of ways to have exactly 2 correct is C(4,2)*C(6,2) = 90, and exactly 3 is C(4,3)*C(6,1)=24. So, total is 115. So, the ratio is 1/115. \\
\textcolor{criticcolor}{<critic>Correct recalculation confirming 115 and ratio.</critic>} \scoreone \textcolor{gray}{[GPT-5-mini: 1]} \\
\addlinespace
\textit{(\dots skipping 13 thinking steps \dots)} \\
\addlinespace
\textcolor{gray}{Step 47.} So, same result. \\
\textcolor{criticcolor}{<critic>Consistent result noted.</critic>} \scoreone \textcolor{gray}{[GPT-5-mini: 1]} \\
\addlinespace
\textcolor{gray}{Step 48.} Therefore, I think that's correct. \\
\textcolor{criticcolor}{<critic>Conclusion is justified.</critic>} \scoreone \textcolor{gray}{[GPT-5-mini: 1]} \\
\addlinespace
\textcolor{gray}{Step 49.} So, the probability is 1/115, so m + n is 1 + 115 = 116. \\
\textcolor{criticcolor}{<critic>Correct restatement of result.</critic>} \scoreone \textcolor{gray}{[GPT-5-mini: 1]} \\
\addlinespace
\textcolor{gray}{Step 50.} \textbf{Final Answer} \boxed{116} \\
</think> \\
\textcolor{criticcolor}{<critic>Final answer is correct.</critic>} \scoreone \textcolor{gray}{[GPT-5-mini: 1]} \\
\addlinespace
\textcolor{gray}{Step 51.} Jen enters a lottery by picking 4 distinct numbers from the set $S = \{1, 2, 3, \ldots, 9, 10\}$. The total number of possible ways to pick her 4 numbers is given by the combination $\binom{10}{4} = 210$. \\
\textcolor{criticcolor}{<critic>Correct restatement of setup and total combinations.</critic>} \scoreone \textcolor{gray}{[GPT-5-mini: 1]} \\
\addlinespace
\textcolor{gray}{Step 52.} To win a prize, she needs at least two of her numbers to be among the four randomly chosen numbers. The probability of winning the grand prize is when all four of her numbers are the randomly chosen numbers. \\
\textcolor{criticcolor}{<critic>Accurate description of winning conditions.</critic>} \scoreone \textcolor{gray}{[GPT-5-mini: 1]} \\
\addlinespace
\textcolor{gray}{Step 53.} First, we calculate the probability of winning the grand prize. There is only one favorable outcome (matching all four numbers), so the probability is: $P(A) = \frac{1}{210}$ \\
\textcolor{criticcolor}{<critic>Correct computation of P(A).</critic>} \scoreone \textcolor{gray}{[GPT-5-mini: 1]} \\
\addlinespace
\textit{(\dots skipping 4 summary blocks \dots)} \\
\addlinespace
\textcolor{gray}{Step 58.} The conditional probability of winning the grand prize given that she won a prize is: $P(A|B) = \frac{P(A)}{P(B)} = \frac{\frac{1}{210}}{\frac{115}{210}} = \frac{1}{115}$ \\
\textcolor{criticcolor}{<critic>Correct application of conditional probability.</critic>} \scoreone \textcolor{gray}{[GPT-5-mini: 1]} \\
\addlinespace
\textcolor{gray}{Step 59.} Therefore, the probability is $\frac{1}{115}$, and the sum of $m$ and $n$ where the probability is $\frac{m}{n}$ is: $m + n = 1 + 115 = 116$ \\
\textcolor{criticcolor}{<critic>Correct final probability and sum.</critic>} \scoreone \textcolor{gray}{[GPT-5-mini: 1]} \\
\addlinespace
\textcolor{gray}{Step 60.} $\boxed{116}$ \\
\textcolor{criticcolor}{<critic>Correct final boxed result.</critic>} \scoreone \textcolor{gray}{[GPT-5-mini: 1]} \\
\addlinespace
\bottomrule
\end{tabularx}
\caption{\textbf{Qualitative example (AIME24, lottery probability).} A typical STC trajectory with paired critiques. At step 31 STC catches the model floating a bogus alternative computation ($\Pr(\text{all four correct} \mid \text{at least two correct}) = 1/\binom{6}{2}$) and notes that the preceding `given at least two' reasoning is the correct path; at steps 32--33 the model returns to the correct enumeration and re-derives the answer $1/115$. The final answer matches the ground truth. Similar to our base model DS-Qwen-1.5B, the post-\texttt{</think>} blocks are STC's brief summary restatement. The gray ``Step $N$'' labels are added for display only; the model emits steps separated by \texttt{\textbackslash n\textbackslash n} without numeric IDs. We also leverage gray \texttt{[GPT-5-mini: $0$/$1$]} tag to show the pseudo-GT judge's verdict.}
\label{tab:qualitative}
\end{table*}

\begin{table*}[!t]
\centering
\tiny
\begin{tabularx}{1\textwidth}{>{\raggedright\arraybackslash}X}
\toprule
\textbf{Stepwise Think-Critique (Full Mode) -- Minerva example} \\
\midrule
\addlinespace

<|User|>: Preamble: A radio interferometer, operating at a wavelength of $1 \mathrm{~cm}$, consists of 100 small dishes, each $1 \mathrm{~m}$ in diameter, distributed randomly within a $1 \mathrm{~km}$ diameter circle. What is the angular resolution of a single dish, in radians? \\
\addlinespace
\textcolor{orange}{Let's think step by step and output the final answer within \textbackslash \textbackslash boxed\{\}.}

\textcolor{orange}{Leverage the output \texttt{\textbackslash n\textbackslash n} as the indicator for separating reasoning steps. Following each reasoning step, please provide a critique and a score, where the critique reasons about the correctness of this step within \texttt{<critic>} and \texttt{</critic>} tags, the score of 1 for denoting correct or 0 denoting incorrect within \texttt{<score>} and \texttt{</score>} tags. Follow the format in this example precisely: [Your thinking process for a step goes here] \texttt{<critic>} [Your brief justification for the score goes here] \texttt{</critic>} \texttt{<score>} [0 or 1] \texttt{</score>}} \\

\arrayrulecolor{gray}\midrule\arrayrulecolor{black}
\addlinespace

<|Assistant|><think> \\
\addlinespace
\textcolor{gray}{Step 1.} Okay, so I have this problem about a radio interferometer and I need to find the angular resolution of a single dish. Hmm, let me start by recalling what angular resolution means. I think it's related to how well the telescope can distinguish between two closely spaced points in the sky. The smal... \\
\textcolor{criticcolor}{<critic>Correct understanding of angular resolution and its purpose.</critic>} \scoreone \textcolor{gray}{[GPT-5-mini: 1]} \\
\addlinespace
\textcolor{gray}{Step 2.} The problem says it's a radio interferometer with 100 small dishes, each 1 meter in diameter, distributed randomly within a 1 km diameter circle. The wavelength operating is 1 cm. Wait, so it's a single dish, not an interferometer. Hmm, okay, so maybe I don't need the interferometer part for this... \\
\textcolor{criticcolor_0}{<critic>Confuses the task; the interferometer data is irrelevant to single-dish resolution.</critic>} \scorezero \textcolor{gray}{[GPT-5-mini: 0]} \\
\addlinespace
\textcolor{gray}{Step 3.} So, the formula for angular resolution in radio telescopes is similar to the optical telescopes, right? It's the diffraction limit, which is given by the wavelength divided by the diameter of the telescope. So, the formula is $\theta$ = $\lambda$ / D, where $\theta$ is the angular resolution in r... \\
\textcolor{criticcolor}{<critic>Correct statement of the Rayleigh formula for resolution.</critic>} \scoreone \textcolor{gray}{[GPT-5-mini: 1]} \\
\addlinespace
\textit{(\dots skipping 12 thinking steps where STC critique matches GT \dots)} \\
\addlinespace
\textcolor{gray}{Step 16.} 1 cm is 0.01 meters, and 1 m is 1 meter. So, 0.01 m / 1 m = 0.01 radians. So, yes, that's correct. \\
\textcolor{criticcolor}{<critic>Correct recheck of units and computation.</critic>} \scoreone \textcolor{gray}{[GPT-5-mini: 1]} \\
\addlinespace
\textcolor{gray}{Step 17.} Alternatively, 1 cm is 0.01 m, so 0.01 m divided by 1 m is 0.01. So, 0.01 radians. \\
\textcolor{criticcolor}{<critic>Correct reaffirmation of the computation.</critic>} \scoreone \textcolor{gray}{[GPT-5-mini: 1]} \\
\addlinespace
\textcolor{gray}{Step 18.} So, I think that's the answer. So, the angular resolution is 0.01 radians. \\
\textcolor{criticcolor}{<critic>Correct conclusion.</critic>} \scoreone \textcolor{gray}{[GPT-5-mini: 1]} \\
\addlinespace
\textcolor{gray}{Step 19.} \textbf{Final Answer} The angular resolution of a single dish is \boxed{0.01} radians. \\
</think> \\
\textcolor{criticcolor}{<critic>Correct final answer.</critic>} \scoreone \textcolor{gray}{[GPT-5-mini: 1]} \\
\addlinespace
\textcolor{gray}{Step 20.} To find the angular resolution of a single dish, we use the formula for angular resolution, which is given by the wavelength divided by the diameter of the telescope. The formula is: \\
\textcolor{criticcolor}{<critic>Accurate description of the method.</critic>} \scoreone \textcolor{gray}{[GPT-5-mini: 1]} \\
\addlinespace
\textcolor{gray}{Step 21.} $\theta = \frac{\lambda}{D}$ \\
\textcolor{criticcolor}{<critic>Correct formula.</critic>} \scoreone \textcolor{gray}{[GPT-5-mini: 1]} \\
\addlinespace
\textcolor{gray}{Step 22.} Given: - Wavelength, $\lambda = 1 \, \text{cm} = 0.01 \, \text{m}$ - Diameter, $D = 1 \, \text{m}$ \\
\textcolor{criticcolor}{<critic>Correct conversion and values.</critic>} \scoreone \textcolor{gray}{[GPT-5-mini: 1]} \\
\addlinespace
\textcolor{gray}{Step 23.} Substituting the given values into the formula: \\
\textcolor{criticcolor}{<critic>Appropriate transition to substitution.</critic>} \scoreone \textcolor{gray}{[GPT-5-mini: 1]} \\
\addlinespace
\textcolor{gray}{Step 24.} $\theta = \frac{0.01 \, \text{m}}{1 \, \text{m}} = 0.01 \, \text{radians}$ \\
\textcolor{criticcolor}{<critic>Correct substitution and result.</critic>} \scoreone \textcolor{gray}{[GPT-5-mini: 1]} \\
\addlinespace
\textcolor{gray}{Step 25.} Thus, the angular resolution of a single dish is: \\
\textcolor{criticcolor}{<critic>Correct lead-in to conclusion.</critic>} \scoreone \textcolor{gray}{[GPT-5-mini: 1]} \\
\addlinespace
\textcolor{gray}{Step 26.} $\boxed{0.01}$ \\
\textcolor{criticcolor}{<critic>Correct final statement of the value.</critic>} \scoreone \textcolor{gray}{[GPT-5-mini: 1]} \\
\addlinespace
\bottomrule
\end{tabularx}
\caption{\textbf{Qualitative example (Minerva, single-dish angular resolution).} STC catches a meta-reasoning error at step 2: the trajectory begins by treating an interferometer description as relevant when the question only asks about a single dish. STC's critic correctly flags this misreading. The final answer matches the ground truth.}
\label{tab:example_minerva_dish}
\end{table*}

\begin{table*}[!t]
\centering
\tiny
\begin{tabularx}{1\textwidth}{>{\raggedright\arraybackslash}X}
\toprule
\textbf{Stepwise Think-Critique (Full Mode) -- MATH-500 example} \\
\midrule
\addlinespace

<|User|>: The Greek army contained two types of soldiers: the upper class and the lower class soldiers. If there were a total of 5 upper class soldiers, and 10 lower class soldiers in a certain part of Athens, and the battle of Thermopylae demands a force of 4 upper class soldiers and 8 lower class soldiers, how many different battalions can be sent? \\
\addlinespace
\textcolor{orange}{Let's think step by step and output the final answer within \textbackslash \textbackslash boxed\{\}.}

\textcolor{orange}{Leverage the output \texttt{\textbackslash n\textbackslash n} as the indicator for separating reasoning steps. Following each reasoning step, please provide a critique and a score, where the critique reasons about the correctness of this step within \texttt{<critic>} and \texttt{</critic>} tags, the score of 1 for denoting correct or 0 denoting incorrect within \texttt{<score>} and \texttt{</score>} tags. Follow the format in this example precisely: [Your thinking process for a step goes here] \texttt{<critic>} [Your brief justification for the score goes here] \texttt{</critic>} \texttt{<score>} [0 or 1] \texttt{</score>}} \\

\arrayrulecolor{gray}\midrule\arrayrulecolor{black}
\addlinespace

<|Assistant|><think> \\
\addlinespace
\textcolor{gray}{Step 1.} Okay, so I need to figure out how many different battalions can be sent based on the given information. Let me read the problem again and break it down step by step. \\
\textcolor{criticcolor}{<critic>Good problem setup; no incorrect claims.</critic>} \scoreone \textcolor{gray}{[GPT-5-mini: 1]} \\
\addlinespace
\textcolor{gray}{Step 2.} First, the problem mentions that there are two types of soldiers in the Greek army: upper class and lower class. In a certain part of Athens, there are 5 upper class soldiers and 10 lower class soldiers. So, that's 5 upper and 10 lower soldiers in total. \\
\textcolor{criticcolor}{<critic>Correct identification of given quantities.</critic>} \scoreone \textcolor{gray}{[GPT-5-mini: 1]} \\
\addlinespace
\textcolor{gray}{Step 3.} Then, the battle of Thermopylae demands a force of 4 upper class soldiers and 8 lower class soldiers. So, the Greek army needs to send a combination of upper and lower class soldiers that meets or exceeds the demand of 4 upper and 8 lower soldiers. \\
\textcolor{criticcolor_0}{<critic>Incorrect requirement; the force must not exceed demand.</critic>} \scorezero \textcolor{gray}{[GPT-5-mini: 0]} \\
\addlinespace
\textcolor{gray}{Step 4.} The question is asking how many different battalions can be sent. I think this means how many different ways we can choose 4 upper class soldiers from the 5 available and 8 lower class soldiers from the 10 available. So, it's a combinatorics problem where we need to calculate the number of combin... \\
\textcolor{criticcolor}{<critic>Correct interpretation of combinations and multiplication principle.</critic>} \scoreone \textcolor{gray}{[GPT-5-mini: 1]} \\
\addlinespace
\textit{(\dots skipping 18 thinking steps where STC critique matches GT \dots)} \\
\addlinespace
\textcolor{gray}{Step 23.} Wait, let me just make sure I didn't make any mistakes here. So, 5 upper class soldiers, choosing 4: that's 5 ways. 10 lower class soldiers, choosing 8: that's 45 ways. Multiply them together to get the total number of combinations, which is 5 * 45 = 225. \\
\textcolor{criticcolor}{<critic>Accurate verification of prior computations.</critic>} \scoreone \textcolor{gray}{[GPT-5-mini: 1]} \\
\addlinespace
\textcolor{gray}{Step 24.} Yes, that seems correct. I don't think I missed anything here. The problem is about combinations, not permutations, since the order of soldiers in a battalion doesn't matter. So, we're just selecting groups, not arranging them in a specific order. \\
\textcolor{criticcolor}{<critic>Correct clarification about combinations vs permutations.</critic>} \scoreone \textcolor{gray}{[GPT-5-mini: 1]} \\
\addlinespace
\textcolor{gray}{Step 25.} Therefore, I think 225 is the correct answer. \\
\textcolor{criticcolor}{<critic>Correct reaffirmation of result.</critic>} \scoreone \textcolor{gray}{[GPT-5-mini: 1]} \\
\addlinespace
\textcolor{gray}{Step 26.} \textbf{Final Answer} The total number of different battalions that can be sent is \boxed{225}. \\
</think> \\
\textcolor{criticcolor}{<critic>Correct final statement of the answer.</critic>} \scoreone \textcolor{gray}{[GPT-5-mini: 1]} \\
\addlinespace
\textcolor{gray}{Step 27.} The problem requires determining the number of different battalions that can be sent given the constraints of the battle of Thermopylae. The Greek army has 5 upper class soldiers and 10 lower class soldiers. The required force is 4 upper class soldiers and 8 lower class soldiers. \\
\textcolor{criticcolor}{<critic>Accurate problem restatement.</critic>} \scoreone \textcolor{gray}{[GPT-5-mini: 1]} \\
\addlinespace
\textcolor{gray}{Step 28.} To solve this, we need to calculate the number of ways to choose 4 upper class soldiers from 5 and 8 lower class soldiers from 10. This is a combinatorics problem involving combinations. \\
\textcolor{criticcolor}{<critic>Correct identification of required combinations.</critic>} \scoreone \textcolor{gray}{[GPT-5-mini: 1]} \\
\addlinespace
\textcolor{gray}{Step 29.} First, we calculate the number of ways to choose 4 upper class soldiers from 5: $C(5, 4) = \frac{5!}{4!(5 - 4)!} = \frac{5!}{4! \cdot 1!} = \frac{120}{24 \cdot 1} = 5$ \\
\textcolor{criticcolor}{<critic>Correct calculation of C(5,4).</critic>} \scoreone \textcolor{gray}{[GPT-5-mini: 1]} \\
\addlinespace
\textcolor{gray}{Step 30.} Next, we calculate the number of ways to choose 8 lower class soldiers from 10. Using the combination formula and the symmetry property of combinations: $C(10, 8) = C(10, 2) = \frac{10!}{2!(10 - 2)!} = \frac{10 \cdot 9 \cdot 8!}{2 \cdot 1 \cdot 8!} = \frac{90}{2} = 45$ \\
\textcolor{criticcolor}{<critic>Correct calculation of C(10,8) via symmetry.</critic>} \scoreone \textcolor{gray}{[GPT-5-mini: 1]} \\
\addlinespace
\textcolor{gray}{Step 31.} Multiplying the number of ways to choose the upper class soldiers by the number of ways to choose the lower class soldiers gives the total number of different battalions: $5 \times 45 = 225$ \\
\textcolor{criticcolor}{<critic>Correct multiplication yielding total combinations.</critic>} \scoreone \textcolor{gray}{[GPT-5-mini: 1]} \\
\addlinespace
\textcolor{gray}{Step 32.} Thus, the total number of different battalions that can be sent is \boxed{225}. \\
\textcolor{criticcolor}{<critic>Correct concluding statement.</critic>} \scoreone \textcolor{gray}{[GPT-5-mini: 1]} \\
\addlinespace
\bottomrule
\end{tabularx}
\caption{\textbf{Qualitative example (MATH-500, Greek army combinations).} At step 3 STC catches a misreading of the problem constraint (the army must field exactly the demanded force, not ``meet or exceed'' it). The final answer matches the ground truth.}
\label{tab:example_math_army}
\end{table*}

\begin{table*}[!t]
\centering
\tiny
\begin{tabularx}{1\textwidth}{>{\raggedright\arraybackslash}X}
\toprule
\textbf{Stepwise Think-Critique (Full Mode) -- OlympiadBench example} \\
\midrule
\addlinespace

<|User|>: For a positive integer $a$, define a sequence of integers $x_{1}, x_{2}, \ldots$ by letting $x_{1}=a$ and $x_{n+1}=2 x_{n}+1$ for $n \geq 1$. Let $y_{n}=2^{x_{n}}-1$. Determine the largest possible $k$ such that, for some positive integer $a$, the numbers $y_{1}, \ldots, y_{k}$ are all prime. \\
\addlinespace
\textcolor{orange}{Let's think step by step and output the final answer within \textbackslash \textbackslash boxed\{\}.}

\textcolor{orange}{Leverage the output \texttt{\textbackslash n\textbackslash n} as the indicator for separating reasoning steps. Following each reasoning step, please provide a critique and a score, where the critique reasons about the correctness of this step within \texttt{<critic>} and \texttt{</critic>} tags, the score of 1 for denoting correct or 0 denoting incorrect within \texttt{<score>} and \texttt{</score>} tags. Follow the format in this example precisely: [Your thinking process for a step goes here] \texttt{<critic>} [Your brief justification for the score goes here] \texttt{</critic>} \texttt{<score>} [0 or 1] \texttt{</score>}} \\

\arrayrulecolor{gray}\midrule\arrayrulecolor{black}
\addlinespace

<|Assistant|><think> \\
\addlinespace
\textcolor{gray}{Step 1.} Alright, so I've got this problem here. Let me try to understand it step by step. It says: \\
\textcolor{criticcolor}{<critic>Introductory setup with no incorrect claims.</critic>} \scoreone \textcolor{gray}{[GPT-5-mini: 1]} \\
\addlinespace
\textcolor{gray}{Step 2.} "For a positive integer $a$, define a sequence of integers $x_{1}, x_{2}, \ldots$ by letting $x_{1} = a$ and $x_{n+1} = 2x_{n} + 1$ for $n \geq 1$. Let $y_{n} = 2^{x_{n}} - 1$. Determine the largest possible $k$ such that, for some positive integer $a$, the numbers $y_{1}, \ldots, y_{k}$ are all... \\
\textcolor{criticcolor}{<critic>Accurate restatement of the problem.</critic>} \scoreone \textcolor{gray}{[GPT-5-mini: 1]} \\
\addlinespace
\textit{(\dots skipping 71 thinking steps where STC critique matches GT \dots)} \\
\addlinespace
\textcolor{gray}{Step 74.} $x_1 = 10$, $y_1 = 2^{10} - 1 = 1024 - 1 = 1023$, which is 3*11*31, composite. \\
\textcolor{criticcolor}{<critic>Correct factorization of 1023.</critic>} \scoreone \textcolor{gray}{[GPT-5-mini: 1]} \\
\addlinespace
\textcolor{gray}{Step 75.} Wait, maybe $a = 1$ is the only one that gives $y_1$ prime, but $y_2$ is not. \\
\textcolor{criticcolor_0}{<critic>Incorrect claim about a = 1 giving only y1 prime.</critic>} \scorezero \textcolor{gray}{[GPT-5-mini: 0]} \\
\addlinespace
\textcolor{gray}{Step 76.} Wait, so $a = 2$ and $a = 3$ give $k = 2$. Let's see if there's an $a$ such that $y_1$ and $y_2$ are both prime. \\
\textcolor{criticcolor}{<critic>Valid idea to look for a with both y1 and y2 prime.</critic>} \scoreone \textcolor{gray}{[GPT-5-mini: 1]} \\
\addlinespace
\textcolor{gray}{Step 77.} So, $y_1 = 2^{a} - 1$ must be prime, and $y_2 = 2^{(a + 1) \cdot 2^{0} - 1} - 1 = 2^{a + 1 - 1} - 1 = 2^{a} - 1$. Wait, no, hold on. \\
\textcolor{criticcolor_0}{<critic>Incorrect assertion about y2 expression.</critic>} \scorezero \textcolor{gray}{[GPT-5-mini: 0]} \\
\addlinespace
\textcolor{gray}{Step 78.} Wait, no, earlier we had $x_n = (a + 1) \cdot 2^{n - 1} - 1$. So, for $n = 2$, $x_2 = (a + 1) \cdot 2^{1} - 1 = 2(a + 1) - 1 = 2a + 2 - 1 = 2a + 1$. So, $y_2 = 2^{2a + 1} - 1$. \\
\textcolor{criticcolor}{<critic>Corrects and computes x2 and y2 correctly.</critic>} \scoreone \textcolor{gray}{[GPT-5-mini: 1]} \\
\addlinespace
\textit{(\dots skipping 98 thinking steps where STC critique matches GT \dots)} \\
\addlinespace
\textcolor{gray}{Step 177.} So, the maximum $k$ is 2, achieved by $a = 2$ and $a = 3$. \\
\textcolor{criticcolor}{<critic>Correct conclusion.</critic>} \scoreone \textcolor{gray}{[GPT-5-mini: 1]} \\
\addlinespace
\textcolor{gray}{Step 178.} Therefore, the answer is $\boxed{2}$. \\
\textcolor{criticcolor}{<critic>Correct final answer.</critic>} \scoreone \textcolor{gray}{[GPT-5-mini: 1]} \\
\addlinespace
\textcolor{gray}{Step 179.} \textbf{Final Answer} \boxed{2} \\
</think> \\
\textcolor{criticcolor}{<critic>Final answer is correct.</critic>} \scoreone \textcolor{gray}{[GPT-5-mini: 1]} \\
\addlinespace
\textcolor{gray}{Step 180.} To determine the largest possible $k$ such that, for some positive integer $a$, the numbers $y_1, \ldots, y_k$ are all prime, we start by analyzing the given sequences. \\
\textcolor{criticcolor}{<critic>Accurate summary of approach.</critic>} \scoreone \textcolor{gray}{[GPT-5-mini: 1]} \\
\addlinespace
\textcolor{gray}{Step 181.} The sequence $x_n$ is defined recursively by $x_1 = a$ and $x_{n+1} = 2x_n + 1$. This sequence can be expressed explicitly as: $x_n = (a + 1) \cdot 2^{n - 1} - 1$ \\
\textcolor{criticcolor}{<critic>Correct explicit formula for x\_n.</critic>} \scoreone \textcolor{gray}{[GPT-5-mini: 1]} \\
\addlinespace
\textcolor{gray}{Step 182.} The sequence $y_n$ is defined as $y_n = 2^{x_n} - 1$. We need to find the largest $k$ such that $y_1, \ldots, y_k$ are all prime. \\
\textcolor{criticcolor}{<critic>Correct definition of y\_n.</critic>} \scoreone \textcolor{gray}{[GPT-5-mini: 1]} \\
\addlinespace
\textcolor{gray}{Step 183.} We compute the first few terms of $x_n$ and $y_n$ for various values of $a$: \\
\textcolor{criticcolor}{<critic>Valid computational approach.</critic>} \scoreone \textcolor{gray}{[GPT-5-mini: 1]} \\
\addlinespace
\textcolor{gray}{Step 184.} - For $a = 2$: - $x_1 = 2$, $y_1 = 2^2 - 1 = 3$ (prime) - $x_2 = 5$, $y_2 = 2^5 - 1 = 31$ (prime) - $x_3 = 11$, $y_3 = 2^{11} - 1 = 2047$ (composite) - Thus, $k = 2$. \\
\textcolor{criticcolor}{<critic>Correct computations for a = 2.</critic>} \scoreone \textcolor{gray}{[GPT-5-mini: 1]} \\
\addlinespace
\textcolor{gray}{Step 185.} - For $a = 3$: - $x_1 = 3$, $y_1 = 2^3 - 1 = 7$ (prime) - $x_2 = 7$, $y_2 = 2^7 - 1 = 127$ (prime) - $x_3 = 15$, $y_3 = 2^{15} - 1 = 32767$ (composite) - Thus, $k = 2$. \\
\textcolor{criticcolor}{<critic>Correct computations for a = 3.</critic>} \scoreone \textcolor{gray}{[GPT-5-mini: 1]} \\
\addlinespace
\textcolor{gray}{Step 186.} - For other values of $a$ (such as 1, 4, 5, 6, 7, 15, etc.), we find that $y_1$ or $y_2$ is composite, making $k$ less than 2. \\
\textcolor{criticcolor}{<critic>Correct observation about other a values.</critic>} \scoreone \textcolor{gray}{[GPT-5-mini: 1]} \\
\addlinespace
\textcolor{gray}{Step 187.} After checking various values, we find that the maximum $k$ is 2, achieved by $a = 2$ and $a = 3$. \\
\textcolor{criticcolor}{<critic>Correct conclusion about maximum k.</critic>} \scoreone \textcolor{gray}{[GPT-5-mini: 1]} \\
\addlinespace
\textcolor{gray}{Step 188.} Thus, the largest possible $k$ is: $\boxed{2}$ \\
\textcolor{criticcolor}{<critic>Correct final statement.</critic>} \scoreone \textcolor{gray}{[GPT-5-mini: 1]} \\
\addlinespace
\bottomrule
\end{tabularx}
\caption{\textbf{Qualitative example (OlympiadBench, prime-$y_n$ sequence).} STC catches two independent errors in the thinking trace: step 75 makes a wrong claim about $a{=}1$ being the only $a$ with $y_1$ prime; step 77 writes an incorrect expression for $y_2$. The final answer matches the ground truth.}
\label{tab:example_olympiad_primes}
\end{table*}

\subsection{Compact Mode}
\label{sec:compact_mode}

Tables~\ref{tab:example-compact} to~\ref{tab:example-compact-olympiad} show the four problems from \S\ref{sec:more_qualitative} re-run in compact mode (system prompt stripped of \texttt{<critic>}/\texttt{<score>} instructions), so the model emits reasoning only. The two modes differ \emph{only} in the system prompt; the underlying STC checkpoint and decoding parameters are identical.

\begin{table*}[!t]
\centering
\tiny
\begin{tabularx}{1\textwidth}{>{\raggedright\arraybackslash}X}
\toprule
\textbf{Stepwise Think-Critique (Compact Mode) -- AIME24 example} \\
\midrule
\addlinespace

<|User|>: Jen enters a lottery by picking $4$ distinct numbers from $S=\{1,2,3,\cdots,9,10\}.$ $4$ numbers are randomly chosen from $S.$ She wins a prize if at least two of her numbers were $2$ of the randomly chosen numbers, and wins the grand prize if all four of her numbers were the randomly chosen numbers. The probability of her winning the grand prize given that she won a prize i... \\
\addlinespace
\textcolor{orange}{Let's think step by step and output the final answer within \textbackslash \textbackslash boxed\{\}.} \\

\arrayrulecolor{gray}\midrule\arrayrulecolor{black}
\addlinespace

<|Assistant|><think> \\
\addlinespace
\textcolor{gray}{Step 1.} Okay, so I've got this probability problem here, and I need to figure it out step by step. Let me try to understand what's being asked. \\
\addlinespace
\textcolor{gray}{Step 2.} First, Jen is entering a lottery by picking 4 distinct numbers from the set S, which is {1,2,3,...,10}. So, S has 10 numbers. Then, 4 numbers are randomly chosen from S. Jen wins a prize if at least two of her numbers are among those 4 chosen. If all four of her numbers are the chosen ones, she wins the grand prize. I need to find the probability that she wins the grand priz... \\
\addlinespace
\textit{(\dots skipping 57 reasoning steps \dots)} \\
\addlinespace
\textcolor{gray}{Step 60.} \textbf{Final Answer}
\boxed{116}
</think> \\
\addlinespace
\textit{(\dots skipping 7 reasoning steps \dots)} \\
\addlinespace
\textcolor{gray}{Step 68.} Thus, the probability is $\frac{1}{115}$, and the sum of $m$ and $n$ where the probability is $\frac{m}{n}$ is $1 + 115 = 116$. \\
\addlinespace
\textcolor{gray}{Step 69.} $\boxed{116}$ \\
\addlinespace
\bottomrule
\end{tabularx}
\caption{\textbf{Compact-mode output (AIME24, lottery probability).} The same problem as Table~\ref{tab:qualitative} but with the critique-instruction stripped from the system prompt. STC outputs reasoning only; the final answer matches the GT.}
\label{tab:example-compact}
\end{table*}

\begin{table*}[!t]
\centering
\tiny
\begin{tabularx}{1\textwidth}{>{\raggedright\arraybackslash}X}
\toprule
\textbf{Stepwise Think-Critique (Compact Mode) -- Minerva example} \\
\midrule
\addlinespace

<|User|>: Preamble: A radio interferometer, operating at a wavelength of $1 \mathrm{~cm}$, consists of 100 small dishes, each $1 \mathrm{~m}$ in diameter, distributed randomly within a $1 \mathrm{~km}$ diameter circle. 

What is the angular resolution of a single dish, in radians? \\
\addlinespace
\textcolor{orange}{Let's think step by step and output the final answer within \textbackslash \textbackslash boxed\{\}.} \\

\arrayrulecolor{gray}\midrule\arrayrulecolor{black}
\addlinespace

<|Assistant|><think> \\
\addlinespace
\textcolor{gray}{Step 1.} Okay, so I need to figure out the angular resolution of a single dish. Hmm, angular resolution... I think that's related to how well the telescope can resolve different sources of light, right? It's like the smallest angle at which two points can be distinguished. I remember something about the diffraction limit, maybe? So, the formula for the diffraction limit of a single s... \\
\addlinespace
\textcolor{gray}{Step 2.} Wait, the formula for the diffraction minimum in single-slit diffraction is given by $a \sin \theta = \lambda$, where $a$ is the width of the slit or the aperture, and $\lambda$ is the wavelength of the light. So, the angular resolution $\theta$ would be $\theta = \frac{\lambda}{a}$. Is that right? Let me confirm. \\
\addlinespace
\textit{(\dots skipping 10 reasoning steps \dots)} \\
\addlinespace
\textcolor{gray}{Step 13.} \textbf{Final Answer}
The angular resolution of a single dish is \boxed{0.01} radians.
</think> \\
\addlinespace
\textit{(\dots skipping 5 reasoning steps \dots)} \\
\addlinespace
\textcolor{gray}{Step 19.} $\theta = \frac{0.01 \text{ m}}{1 \text{ m}} = 0.01 \text{ radians}$ \\
\addlinespace
\textcolor{gray}{Step 20.} Thus, the angular resolution of a single dish is $\boxed{0.01}$ radians. \\
\addlinespace
\bottomrule
\end{tabularx}
\caption{\textbf{Compact-mode output (Minerva, single-dish angular resolution).} Counterpart of Table~\ref{tab:example_minerva_dish}.}
\label{tab:example-compact-minerva}
\end{table*}

\begin{table*}[!t]
\centering
\tiny
\begin{tabularx}{1\textwidth}{>{\raggedright\arraybackslash}X}
\toprule
\textbf{Stepwise Think-Critique (Compact Mode) -- MATH-500 example} \\
\midrule
\addlinespace

<|User|>: The Greek army contained two types of soldiers: the upper class and the lower class soldiers. If there were a total of 5 upper class soldiers, and 10 lower class soldiers in a certain part of Athens, and the battle of Thermopylae demands a force of 4 upper class soldiers and 8 lower class soldiers, how many different battalions can be sent? \\
\addlinespace
\textcolor{orange}{Let's think step by step and output the final answer within \textbackslash \textbackslash boxed\{\}.} \\

\arrayrulecolor{gray}\midrule\arrayrulecolor{black}
\addlinespace

<|Assistant|><think> \\
\addlinespace
\textcolor{gray}{Step 1.} Alright, so I have this problem about the Greek army and different types of soldiers. Let me try to figure it out step by step. Maybe I'll write down my thoughts as I go along. \\
\addlinespace
\textcolor{gray}{Step 2.} First, the problem says that the Greek army has two types of soldiers: upper class and lower class. There are 5 upper class soldiers and 10 lower class soldiers in a certain part of Athens. Then, the battle of Thermopylae demands a force of 4 upper class soldiers and 8 lower class soldiers. The question is asking how many different battalions can be sent. \\
\addlinespace
\textit{(\dots skipping 29 reasoning steps \dots)} \\
\addlinespace
\textcolor{gray}{Step 32.} \textbf{Final Answer}
The number of different battalions that can be sent is \boxed{225}.
</think> \\
\addlinespace
\textit{(\dots skipping 4 reasoning steps \dots)} \\
\addlinespace
\textcolor{gray}{Step 37.} Thus, the number of different battalions that can be sent is $\boxed{225}$. \\
\addlinespace
\textcolor{gray}{Step 38.} $\boxed{225}$ \\
\addlinespace
\bottomrule
\end{tabularx}
\caption{\textbf{Compact-mode output (MATH-500, Greek army combinations).} Counterpart of Table~\ref{tab:example_math_army}.}
\label{tab:example-compact-math}
\end{table*}

\begin{table*}[!t]
\centering
\tiny
\begin{tabularx}{1\textwidth}{>{\raggedright\arraybackslash}X}
\toprule
\textbf{Stepwise Think-Critique (Compact Mode) -- OlympiadBench example} \\
\midrule
\addlinespace

<|User|>: For a positive integer $a$, define a sequence of integers $x_{1}, x_{2}, \ldots$ by letting $x_{1}=a$ and $x_{n+1}=2 x_{n}+1$ for $n \geq 1$. Let $y_{n}=2^{x_{n}}-1$. Determine the largest possible $k$ such that, for some positive integer $a$, the numbers $y_{1}, \ldots, y_{k}$ are all prime. \\
\addlinespace
\textcolor{orange}{Let's think step by step and output the final answer within \textbackslash \textbackslash boxed\{\}.} \\

\arrayrulecolor{gray}\midrule\arrayrulecolor{black}
\addlinespace

<|Assistant|><think> \\
\addlinespace
\textcolor{gray}{Step 1.} Okay, so I have this problem here where I need to find the largest possible k such that for some positive integer a, the numbers y?, y?, ..., y? are all prime. The definitions given are x? = a, and each subsequent term is x\_{n+1} = 2x? + 1. Then y? is defined as 2\^{}{x?} - 1. \\
\addlinespace
\textcolor{gray}{Step 2.} Hmm, let me try to break this down step by step. First, let's understand the sequence x?. It starts with some positive integer a, and each term is twice the previous term plus one. So, it's a recurrence relation. \\
\addlinespace
\textit{(\dots skipping 123 reasoning steps \dots)} \\
\addlinespace
\textcolor{gray}{Step 126.} \textbf{Final Answer}
The largest possible $k$ is \boxed{2}.
</think> \\
\addlinespace
\textit{(\dots skipping 7 reasoning steps \dots)} \\
\addlinespace
\textcolor{gray}{Step 134.} Thus, the largest possible $k$ such that $y_1, y_2, \ldots, y_k$ are all prime is $k = 2$. \\
\addlinespace
\textcolor{gray}{Step 135.} $\boxed{2}$ \\
\addlinespace
\bottomrule
\end{tabularx}
\caption{\textbf{Compact-mode output (OlympiadBench, prime-$y_n$ sequence).} Counterpart of Table~\ref{tab:example_olympiad_primes}.}
\label{tab:example-compact-olympiad}
\end{table*}

\subsection{Failure Case}
\label{sec:failure_cases}

Following \S\ref{subsub:qual}, Table~\ref{tab:wrong_example_minerva} shows a failure case where STC's critic itself is unreliable: at step 3 it confirms a wrong unit conversion (``3 Mpc equals 3000 parsecs'') with \scoreone. The pseudo-GT GPT-5-mini judge identifies ``\textit{Incorrect unit conversion: $1\,\mathrm{Mpc}=10^{6}\,\mathrm{pc}$, so $3\,\mathrm{Mpc}=3{,}000{,}000\,\mathrm{pc}$, not $3{,}000\,\mathrm{pc}$.}'' This example shows that, while STC's natural-language critique offers step-level interpretability, the critic itself is not 100\% reliable; further improving its performance remains a direction for future work.

\begin{table*}[!t]
\centering
\tiny
\begin{tabularx}{1\textwidth}{>{\raggedright\arraybackslash}X}
\toprule
\textbf{Stepwise Think-Critique (Full Mode) -- Minerva wrong-answer example (critic under-flags)} \\
\midrule
\addlinespace

<|User|>: Subproblem 0: A particular star has an absolute magnitude $M=-7$. If this star is observed in a galaxy that is at a distance of $3 \mathrm{Mpc}$, what will its apparent magnitude be? Solution: $\text { Given: } M=-7 \text { and } d=3 \mathrm{Mpc}$... \\
\addlinespace
\textcolor{orange}{Let's think step by step and output the final answer within \textbackslash \textbackslash boxed\{\}.}

\textcolor{orange}{Leverage the output \texttt{\textbackslash n\textbackslash n} as the indicator for separating reasoning steps. Following each reasoning step, please provide a critique and a score, where the critique reasons about the correctness of this step within \texttt{<critic>} and \texttt{</critic>} tags, the score of 1 for denoting correct or 0 denoting incorrect within \texttt{<score>} and \texttt{</score>} tags. Follow the format in this example precisely: [Your thinking process for a step goes here] \texttt{<critic>} [Your brief justification for the score goes here] \texttt{</critic>} \texttt{<score>} [0 or 1] \texttt{</score>}} \\

\arrayrulecolor{gray}\midrule\arrayrulecolor{black}
\addlinespace

<|Assistant|><think> \\
\addlinespace
\textcolor{gray}{Step 1.} Alright, so I have two subproblems here about magnitudes and distances. Let me tackle them one by one. \\
\textcolor{criticcolor}{<critic>Clear organization of tasks; no mathematical errors.</critic>} \scoreone \textcolor{gray}{[GPT-5-mini: 1]} \\
\addlinespace
\textcolor{gray}{Step 2.} Starting with Subproblem 0: The star has an absolute magnitude of M = -7, and it's observed in a galaxy that's 3 Mpc away. I need to find its apparent magnitude. Hmm, okay. I remember that apparent magnitude is how bright the star appears from Earth, while absolute magnitude i... \\
\textcolor{criticcolor}{<critic>Correct statement of the formula and units.</critic>} \scoreone \textcolor{gray}{[GPT-5-mini: 1]} \\
\addlinespace
\textcolor{gray}{Step 3.} So, let me plug in the numbers. The distance is given as 3 Mpc. Wait, Mpc stands for megaparsecs, right? So, 3 Mpc is 3,000 parsecs. Got it. So, d is 3,000 parsecs. \\
\textcolor{criticcolor}{<critic>Correct conversion: 3 Mpc equals 3000 parsecs.</critic>} \scoreone \textcolor{gray}{[GPT-5-mini: 0]} \\
\addlinespace
\textcolor{gray}{Step 4.} Apparent magnitude m = -7 + 5 * log(d/10). Let me compute d/10 first. That's 3,000 / 10 = 300. So, log(300). Hmm, I need to remember what log base 10 of 300 is. \\
\textcolor{criticcolor}{<critic>Correct substitution and computation of d/10.</critic>} \scoreone \textcolor{gray}{[GPT-5-mini: 0]} \\
\addlinespace
\textcolor{gray}{Step 5.} I know that log(100) is 2, and log(300) is log(3*100) = log(3) + log(100). Log(3) is approximately 0.4771, so log(300) $\approx$ 0.4771 + 2 = 2.4771. \\
\textcolor{criticcolor}{<critic>Correct use of log properties and value for log 3.</critic>} \scoreone \textcolor{gray}{[GPT-5-mini: 1]} \\
\addlinespace
\textit{(\dots skipping 49 thinking steps \dots)} \\
\addlinespace
\textcolor{gray}{Step 55.} Therefore, I think the correct answers are approximately 5.39 for Subproblem 0 and 12.39 for Subproblem 1. \\
\textcolor{criticcolor}{<critic>Correct restatement of final answers.</critic>} \scoreone \textcolor{gray}{[GPT-5-mini: 0]} \\
\addlinespace
\textcolor{gray}{Step 56.} \textbf{Final Answer} \\
\textcolor{criticcolor}{<critic>Section header; no mathematical claim.</critic>} \scoreone \textcolor{gray}{[GPT-5-mini: 1]} \\
\addlinespace
\textcolor{gray}{Step 57.} The final answer for Subproblem 0 is \boxed{5.39} and for Subproblem 1 is \boxed{12.39}. \\
</think> \\
\textcolor{criticcolor}{<critic>Correct final numeric answers.</critic>} \scoreone \textcolor{gray}{[GPT-5-mini: 0]} \\
\addlinespace
\textcolor{gray}{Step 58.} Given the absolute magnitude $M = -7$ and the distance $d = 3 \, \text{Mpc}$: \\
\textcolor{criticcolor}{<critic>Accurate restatement of given values.</critic>} \scoreone \textcolor{gray}{[GPT-5-mini: 1]} \\
\addlinespace
\textcolor{gray}{Step 59.} 1. \textbf{Apparent Magnitude Calculation:} - Convert the distance from Mpc to parsecs: $3 \, \text{Mpc} = 3000 \, \text{pc}$. - Use the formula for apparent magnitude: $m = M + 5 \log_{10}\left(\frac{d}{10 \, \text{pc}}\right)$ - Substitute the values:... \\
\textcolor{criticcolor}{<critic>Correct, detailed derivation and computation yielding 5.39.</critic>} \scoreone \textcolor{gray}{[GPT-5-mini: 0]} \\
\addlinespace
\textit{(\dots skipping 1 summary blocks \dots)} \\
\addlinespace
\textcolor{gray}{Step 61.} \#\#\# Final Answer \\
\textcolor{criticcolor}{<critic>Section header; no mathematical content.</critic>} \scoreone \textcolor{gray}{[GPT-5-mini: 1]} \\
\addlinespace
\textcolor{gray}{Step 62.} The final answer for Subproblem 0 is $\boxed{5.39}$ and for Subproblem 1 is $\boxed{12.39}$. \\
\textcolor{criticcolor}{<critic>Correct final answers restated.</critic>} \scoreone \textcolor{gray}{[GPT-5-mini: 0]} \\
\addlinespace
\bottomrule
\end{tabularx}
\caption{\textbf{Failure example (Minerva, apparent-magnitude formula)} -- \emph{critic under-flags}. At step 3 the model claims ``3 Mpc equals 3000 parsecs'' (the correct value is $3 \times 10^6$ parsecs); STC's critic confirms this with \scoreone (``Correct conversion''), establishing a wrong distance that propagates through the entire trajectory. The pseudo-GT GPT-5-mini judge instead flags this with $0$, reasoning: ``\textit{Incorrect unit conversion: $1\,\mathrm{Mpc} = 10^6\,\mathrm{pc}$, so $3\,\mathrm{Mpc} = 3{,}000{,}000\,\mathrm{pc}$, not $3{,}000\,\mathrm{pc}$.}'' GPT-5-mini further flags steps 4 onward as incorrect because the model continues to apply the wrong distance (``\textit{Arithmetic using the incorrect distance is based on the wrong conversion}''), and the final boxed value as inconsistent with the correct distance modulus $\approx 27.39$. The model arrives at the wrong final answer 12.39 (ground truth 27.39).}
\label{tab:wrong_example_minerva}
\end{table*}

\section{Additional Discussion}

\subsection{Computation Cost}
We validate STC on a 1.5B model. We set the maximum response length as 8,192 tokens for reasoning and 4,096 tokens for critique. Running 1200 RL steps requires approximately 5 days on 16$\times$40GB A100 GPUs. 

Our goal is to establish the core mechanism; the framework is model-agnostic and extendable to other models. Due to limited resources, we did not extend the validation to larger scale or multimodal experiments. We hope this work motivates further research on critical-thinking mechanisms.

\subsection{About Reward Hacking} 

One may wonder whether our scheme faces reward hacking. Actually, we did not observe evidence of reward hacking in extensive inspection. 

First, as shown in Table~\ref{tab:answer_level_confusion}, STC identifies 21\% of final answers as incorrect, rather than collapsing to predicting all answers as either correct or incorrect. Similarly, STC identifies 23\% of steps as incorrect on average, close to the pseudo ground-truth ratio of 19.1\% (see Table~\ref{tab:process_level_confusion}).

Second, as shown in Table~\ref{tab:qualitative}, our critic correctly flags a bogus alternative computation that the model floats mid-trace, providing valuable information rather than merely stylistic output.

\subsection{Inference Cost}

The full mode interleaves critiques with reasoning and consumes additional tokens at inference time. Naive deployment without exposing the compact-mode fallback could increase cost, latency, and energy use in user-facing applications. We recommend defaulting to compact mode and enabling full mode only when interpretability is required.

% ===== Ethical Considerations section moved to main body as §7 =====
% \section{Ethical Considerations and Potential Risks}
%
% STC is a methodology study evaluated on standard public mathematical reasoning benchmarks; it does not introduce new user-facing systems, sensitive datasets, or task domains beyond those already established in the LLM reasoning literature. Nevertheless, we identify two concrete risks that warrant disclosure.
%
% \noindent\textbf{Critique reliability.} As shown in Table~\ref{tab:ans_critic_results_wo_VR} of the main manuscript, STC's error-detection accuracy on the answer level remains well below 100\% (averaging 42.6\%); a positive critique should therefore not be interpreted as a correctness guarantee, particularly in high-stakes settings such as education, tutoring, or scientific computing where users may over-trust an explicit score-$1$ signal.
%
% \noindent\textbf{Inference cost.} (moved to D.3 Inference Cost)
%
% We do not foresee misuse pathways unique to STC beyond those already common to LLM-based mathematical assistants.

\end{document}